\documentclass{article}

\usepackage{arxiv}

\usepackage[utf8]{inputenc}
\usepackage[T1]{fontenc}
\usepackage{graphicx,verbatim}
\usepackage{url}
\usepackage{array}
\usepackage{amsmath}
\usepackage[table]{xcolor}
\usepackage[figuresright]{rotating}
\usepackage{atbegshi}
\usepackage{refcount}
\usepackage[numbers,sort&compress]{natbib}
\AtBeginShipout{%
  \ifdefined\pdfpageattr
    \ifnum\value{page}=\getpagerefnumber{tab:capabilities}\relax
      \pdfpageattr{/Rotate 90}%
    \else
      \ifnum\value{page}=\getpagerefnumber{tab:health}\relax
        \pdfpageattr{/Rotate 90}%
      \else
        \pdfpageattr{}%
      \fi
    \fi
  \fi}
\newcommand{\rp}[1]{\href{https://github.com/#1}{{\scriptsize\ttfamily #1}}}
\newcommand{\rel}[2]{#1\newline\mbox{#2}}
\newcommand{\capfull}{$\bullet$}
\newcommand{\cappart}{$\circ$}
\newcommand{\capnone}{x}
\newcommand{\capunclear}{---}
\newcommand{\capna}{\cellcolor{gray!15}}
\newcommand{\capclaim}{\textsuperscript{$\star$}}

\newcommand{\flipversion}{\textbf{v0.10.0}}
\newcommand{\flipreleasedate}{29~September 2026}

\usepackage[hidelinks]{hyperref}
\usepackage{authblk}

\hypersetup{%
  pdftitle={Making Cross-Continental Federated Learning Repeatable with FLIP: a Multi-Application Study},
  pdfauthor={Rafael Garcia-Dias and Alexandre Triay Bagur and Chayanin Tangwiriyasakul and others},
  pdfkeywords={federated learning, medical imaging, platform, governance, deployment}}
\begin{document}
\title{Making Cross-Continental Federated Learning Repeatable with FLIP: a Multi-Application Study}
\renewcommand{\shorttitle}{Cross-Continental Federated Learning with FLIP}
%
\author[1,$\star$]{Rafael Garcia-Dias\thanks{Corresponding author: \texttt{rafael.garcia-dias@kcl.ac.uk}}}
\author[1,$\star$]{Alexandre Triay Bagur}
\author[1,$\star$]{Chayanin Tangwiriyasakul}
\author[1]{Virginia Fernandez}
\author[1]{Parhom Esmaeili}
\author[2]{Piyalitt Ittichaiwong}
\author[1]{Yang Li}
\author[1]{Lawrence Adams}
\author[2]{Wason Buncharoen}
\author[1]{Martin Chapman}
\author[2]{Benjamaporn Chayanond}
\author[2]{Sadthavud Chunrod}
\author[2]{Tanawat Fongsri}
\author[1]{Kass Gibson}
\author[2]{Supat Plungprasertkul}
\author[2]{Supawit Tangpanithandee}
\author[2]{Kanyakorn Veerakanjana}
\author[1]{Vicky Goh}
\author[1]{Michela Antonelli}
\author[1]{Joe Zhang}
\author[2]{Kongkiat Kespechara}
\author[1]{Sebastien Ourselin}
\author[1]{M. Jorge Cardoso}
\affil[1]{School of Biomedical Engineering \& Imaging Sciences, King's College London, London, UK}
\affil[2]{Bangkok Dusit Medical Services, Bangkok, Thailand}
\affil[$\star$]{These authors contributed equally to this work.}
\date{}
\maketitle

\begin{abstract}
	Federated learning (FL) in healthcare remains challenging, as the overhead of rebuilding governance guarantees for every collaboration stops most projects at the proof-of-concept stage. Here we present FLIP (Federated Learning Interoperability Platform), an open-source, multi-application platform that makes FL training and evaluation repeatable. FLIP implements common FL workflows as a set of composable services: cohort queries against per-site structured databases, on-demand DICOM retrieval from institutional PACS, per-site project approval, and reusable FL job types. To demonstrate FLIP, we ran two distinct use cases, federated fine-tuning and federated evaluation, on synthetic chest X-ray cohorts across two client nodes based in the United Kingdom (UK) and Thailand. In FLIP, each institution independently approves its participation in each project and operates its own node under local IT security processes. This study makes an operational rather than an algorithmic claim. It does not compare federated with centralised training; for that question, we refer the reader to existing systematic reviews and meta-analyses. The central result is evidence that such platforms enable international FL collaboration and improve repeatability, auditability, and site-specific governance. We also present a comprehensive comparison of existing platforms to help researchers and operators choose the right platform for their use case.
\end{abstract}

\section{Introduction}
\label{sec:intro}
AI in healthcare depends on access to representative data, yet such data are distributed across hospitals and health systems. Moreover, working with patient data requires privacy protection, governance processes and ethics approvals, and is subject to restrictions on data transfer~\cite{Cardoso2020npjFed}. Federated learning (FL) addresses this by allowing sites to train or evaluate shared models through secure parameter exchange while data remain local~\cite{sheller2020}. However, real-world deployment remains difficult because institutions differ in infrastructure, security requirements, compute resources, data curation, and governance processes~\cite{Bujotzek2024RealWorldRadiology,Dayan2021FLcovidlarge}, and evaluation frameworks for clinical deployment remain underdeveloped~\cite{degrave2025datavalue}. The resulting engineering and governance burden means that each new collaboration typically recreates similar infrastructure~\cite{Michaelides2025,Zhang2024}. As a consequence, academic--industry healthcare collaborations remain rare or limited to research projects with a single scope~\cite{Kuo2025CrossSiloPractice,Zhang2024}. Secure, reusable infrastructure with robust governance mechanisms is therefore needed~\cite{Eden2025}.

\begin{figure}[ht]
	\includegraphics[width=\textwidth]{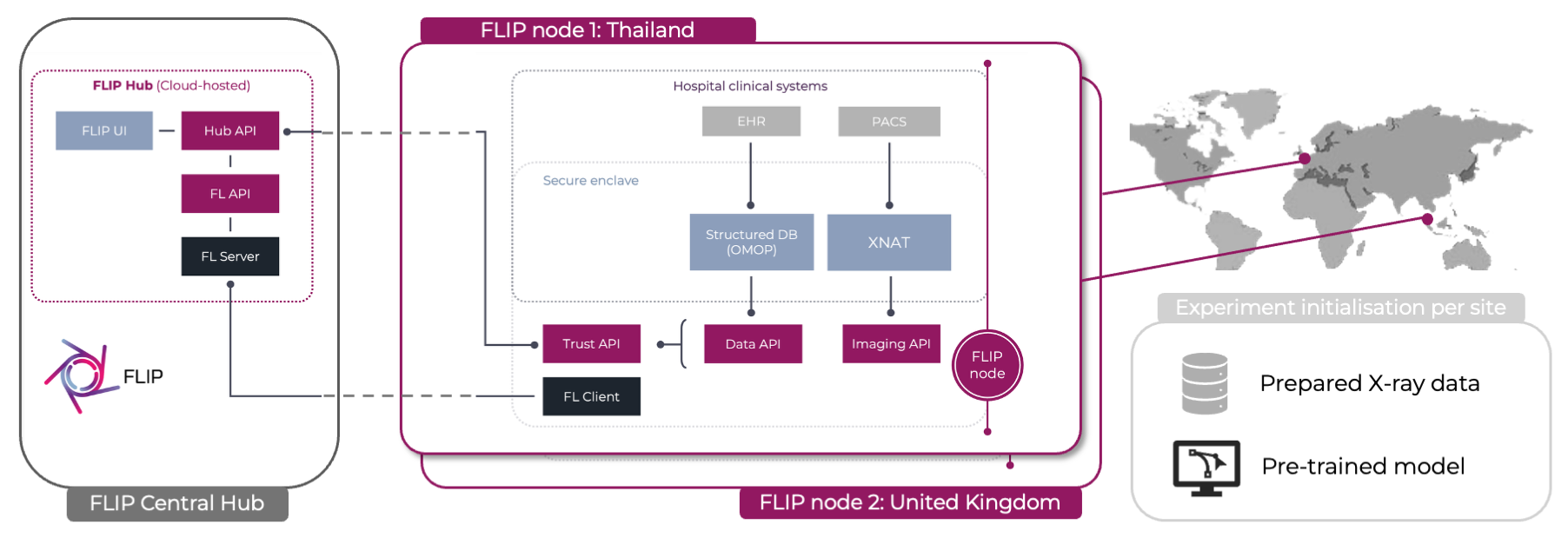}
	\centering
	\caption{FLIP architecture. The Central Hub (left, cloud-hosted) handles coordination and the FL server, while each Trust Node (middle) hosts an OMOP database, XNAT, and an FL client attached to GPU compute. Data leaving the TN are limited to model weight updates, aggregate cohort counts, per-round scalar metrics, and execution logs; no row-level records or image data are transferred.}
	\label{flip_diagram}
\end{figure}

FLIP (Federated Learning Interoperability Platform; \url{github.com/londonaicentre/FLIP}) is a multi-application open-source FL platform (Apache~2.0) that enables repeatable federated AI workflows, including those in medical imaging. It operationalises the lifecycle of a federated project, from cohort discovery and staging to local execution, aggregation over authenticated TLS channels, and auditable provenance. It does so through modular microservices organised into a Central Hub for coordination and one Trust Node per site for local execution (Fig.~\ref{flip_diagram}). FLIP is not restricted to a single imaging modality or FL project. It has recently been used for population-scale phenotyping~\cite{Garcia-Dias2025oldFLIP}, and in this study we demonstrate two additional FL applications (federated fine-tuning and federated model evaluation) for chest X-ray (CXR) classification.

\textbf{Scope of this work.} FLIP is a governance and integration platform, and this paper evaluates it as such. We do not set out to show that federated learning is more accurate than centralised training, and readers looking for that evidence should turn to the aggregate literature rather than to this study. This literature is now large enough to answer the question in terms of effect sizes. A systematic review of 160 articles, covering 710 decentralised models and 8{,}149 head-to-head performance comparisons, finds centralised training favoured in 78\% of comparisons on threshold-dependent metrics such as accuracy and Dice, but in only 51\% on ranking metrics such as AUROC, and then with small effect sizes. Where both approaches reached clinical viability, the median difference was 0.7--1.5 percentage points~\cite{Diniz2026Decentralized}. A meta-analysis of federated mortality prediction reports a pooled AUC of 0.81 (95\% CI 0.76--0.85) for federated models against 0.82 (0.77--0.86) for centralised ones~\cite{Tahir2025FLMortalityMeta}, and reviews specific to medical imaging reach the same qualitative conclusion~\cite{Ghosh2026FLImagingReview,Li2025FLPitfalls}. Federation therefore carries a real but small cost, which is concentrated in operating-point-dependent metrics and amplified by data heterogeneity~\cite{noniidsurvey2024,OgierDuTerrail2022FLamby}.

We adopt federated learning despite this cost. In practice, the comparison that decides whether a multi-institutional study happens at all is seldom federated against centralised training, because centralisation is frequently not a lawful or contractually available option. Instead, the relevant comparison is between a federated study and what a single institution can do alone. For this comparison, the same review finds decentralised models consistently ahead of locally trained ones (86\% favourability on precision and 83\% on accuracy), with median gains of 7.6--27 percentage points where local training had fallen short of clinical viability~\cite{Diniz2026Decentralized}. This is consistent with the large multi-site studies that motivated the field~\cite{sheller2020,Dayan2021FLcovidlarge,Pati2022FeTS}. What blocks such studies in practice is not the aggregation algorithm but the work of connecting sites and satisfying the governance requirements of each one~\cite{Eden2025,Li2025FLPitfalls,Michaelides2025}. FLIP addresses this problem, and the claims made here are therefore operational. The AUROC values we report establish that a real cross-continental federation executed a complete workflow end to end. They do not show that federation improved on a centralised alternative, which we did not run.

Prior medical-imaging FL studies demonstrate that technical feasibility alone does not guarantee uniform benefit across heterogeneous sites~\cite{sheller2020}, and privacy and security considerations mean that deployment conditions matter as much as algorithms~\cite{kaissis2020}. FLIP addresses this gap by providing a reusable infrastructure layer that securely isolates governance, data access, and FL coordination from researchers' per-project algorithm development. Our contributions are threefold: (1) we demonstrate FLIP's general-purpose design by running two distinct FL application types across two sites based in the UK and Thailand, with minimal platform overhead relative to training time; (2) we present a case study from the researcher's perspective, showing that once a site is onboarded, FLIP reduces the marginal per-project engineering to writing one Structured Query Language (SQL) query and uploading the application files; and (3) we describe the cohort-query and job-submission interfaces that make FLIP extensible to further use cases, and delimit what this study does and does not establish about that extensibility.

\section{FLIP: a Multi-Application Platform}

\textbf{Architecture.} The \textbf{Central Hub (CH)} provides the web UI and API for project definition, cohort queries, application upload, role-based access control, and observability. It also hosts object storage for FL application bundles, model checkpoints, and results. FL services consist of independent \emph{networks} or \emph{federations}, each with one FL server plus one client per site, and a CH-side scheduler that dispatches each approved job. FLIP supports NVIDIA FLARE~\cite{Roth2022} and Flower~\cite{flower2022}, both exposing reusable job types whose fixed server-side design isolates researchers' FL apps from the coordination logic. Each \textbf{Trust Node (TN)} runs the same microservices stack inside the local institutional environment: a trust API orchestrates approved work and delegates to a data API for OMOP cohort queries~\cite{OHDSI}, and to an imaging API for DICOM retrieval from PACS into XNAT~\cite{Marcus2007}. The TN FL client attaches local GPU compute. Researchers' FL applications (also known as `apps' or `bundles') access the resolved cohort (dataframe and images) through a dedicated FLIP Python library rather than calling site services directly. TN-to-CH communication is outbound-only over HTTPS (TNs poll for approved tasks, avoiding inbound firewall rules), and images never leave the TN~\cite{kaissis2020}. In this study, the CH ran on AWS ECS Fargate, the UK TN on a desktop RTX~5090, and the Thailand TN on an NVIDIA~L4.

\subsection{Extensible by Design}
\label{subsec:generalPurpose}

FLIP's architecture supports a range of FL workflows in medical imaging through three design choices, described below.

\begin{figure}[ht]
	\centering
	\includegraphics[width=\textwidth]{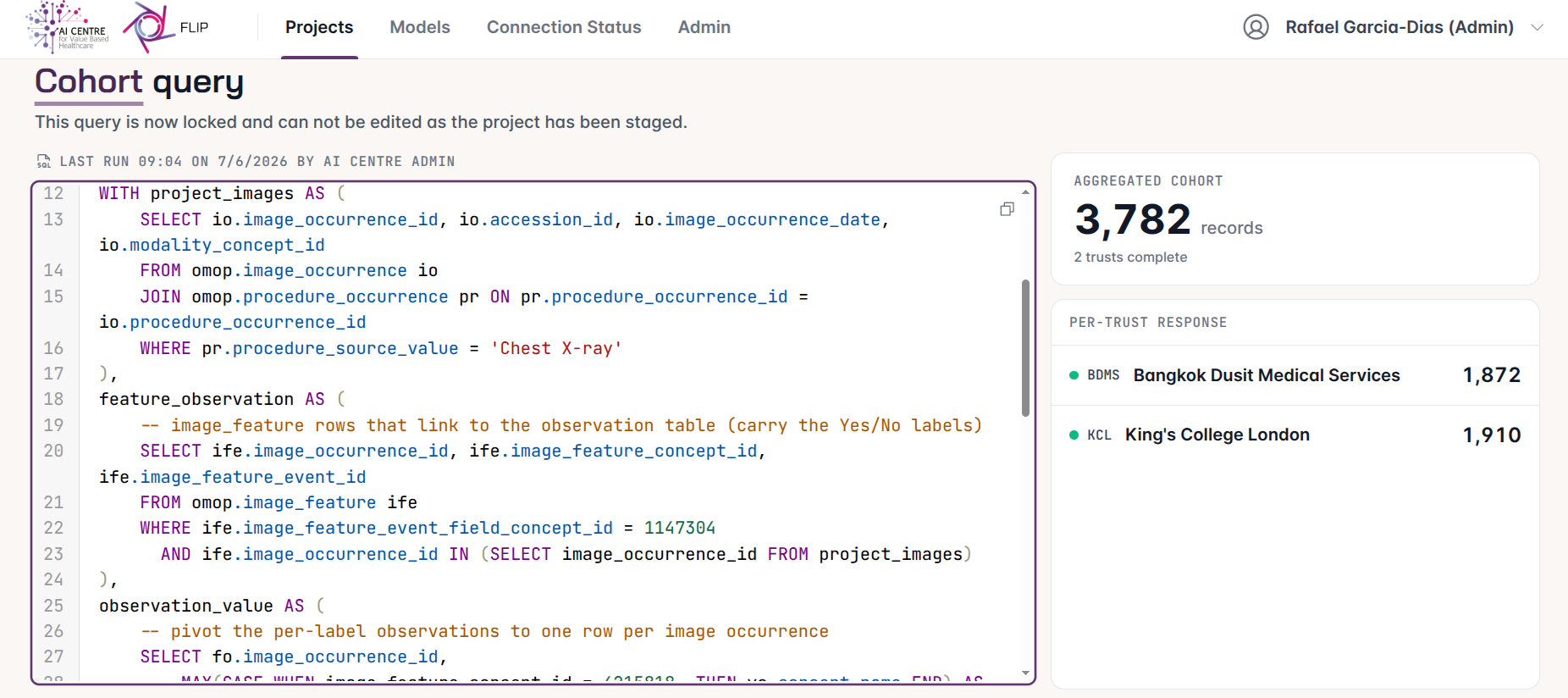}
	\caption{Screenshot of the cohort query for the training set, with the aggregated per-site record counts on the right. A \texttt{WHERE} clause separates the training and holdout sets. The platform accepts SQL \texttt{SELECT} statements only; the full query is available in the FLIP repository.}
	\label{cohortsql}
\end{figure}

\textbf{Flexible cohort queries.} Cohort selection is written as a SQL query that is executed against each site's local OMOP database. Any cohort definable in OMOP (by procedure, diagnosis, age, or laboratory values) is supported. Each site presents OMOP CDM 5.4 extended with the medical-imaging tables of Park et al.~\cite{park2024development} (a distinct extension that shares the name MI-CDM~\cite{Kalokyri2023MICDM}), which allows a cohort to be filtered on imaging parameters as well as clinical ones. FLIP adds an \texttt{accession\_id} column to \texttt{image\_occurrence} so that a resolved cohort joins directly to the studies to be retrieved. Figure~\ref{cohortsql} shows an excerpt of the query used to retrieve the CXR training set in this study; the full query, including the OMOP concept-level pivot for pathology labels, is available in our code repository.

\textbf{Reusable job types.} FLIP implements a number of reusable job types, including supervised training (FedAvg~\cite{McMahan2016} with configurable privacy filters and checkpoint staging) and federated evaluation (single-model or multi-model with automated DeLong significance reporting). Each job type defines a fixed server-side contract; user-submitted FL apps implement only the client-side training or evaluation workloads. New job types (e.g., image synthesis, image translation) require only registering a new server-side configuration.

\textbf{Application plug and play.} Because FL apps interact with the platform only through a job type's fixed client-side contract and the FLIP cohort-access library (never calling site services directly), a functionally new application can be deployed without changing any CH or TN code, since only the new app bundle needs to be uploaded and the corresponding job type selected. We demonstrate this by deploying two functionally unrelated applications, federated fine-tuning and federated evaluation, on identical infrastructure and cohorts with no platform reconfiguration between them (Section~\ref{sec:study}).

The two applications exercised in this study are not two settings of the same
workflow. Federated training is iterative: the server broadcasts a global model,
each site trains it locally on its own data, returns a weight update, and the
server aggregates the updates into a new global model once per round. For
evaluation, the platform sends a model to each site, which scores it against its
own holdout set, reports per-class metrics (AUROC, together with pairwise DeLong
tests across the evaluated models), and returns no weight update, so no site can
influence any model through this workflow. FLIP implements the two as separate
job types, one driving federated rounds and one driving a single cross-site
scoring pass, in both the FLARE and the Flower backend. A site that approves a
project therefore sees which of the two it is running before it runs, and the
distinction is not left to convention in the application code. The evaluation
job type compares several models in a single run, and we used it that way in
this study by scoring the pretrained and the fine-tuned model side by side at
each site (Section~\ref{sec:study}). In FLARE terms, the first application sits
in the ``ScatterAndGather'' workflow and the second in its counterpart for
cross-site model evaluation.

Beyond this study, the FLIP repository ships a public catalogue of example
applications at
\url{https://github.com/londonaicentre/FLIP/tree/develop/fl-tutorials}, covering
image segmentation, image classification, model evaluation, image synthesis with
a diffusion model, and EHR risk prediction from tabular OMOP data alone, with no
imaging path at all. These are complete FL applications, each with a cohort
query, preprocessing chain, model definition and job-type manifest, packaged as
a bundle that is uploaded through the portal exactly as ours were, and they
cover both of FLIP's backends. What we claim for them is what the repository
enforces. The catalogue carries a CPU-only test suite that runs in continuous
integration on every change to it, pinning each application's preprocessing
chain against the raw DICOM pixel data, and checking that every application
declares a job type that a server-side template implements and ships every file
that job type requires. This check exists because the failure it prevents is
silent: a transposed preprocessing chain once fed the foundation model sideways
radiographs, depressing every reported metric without raising an error. Each
example is also exercised end to end on the FLARE simulator through the tutorial
runner, and the full platform path, from portal upload to results, through an
end-to-end smoke test against a running deployment. Since release~\flipversion\ (\flipreleasedate),
both are release gates: the whole catalogue is run on both backends before a
release is published, so the state of the examples at a given version is a
checked property of that release rather than a claim about the source tree. The catalogue is therefore
the evidence for the architectural claim of this section. The job types and the
app contract carry applications well beyond the two deployed in the study, and
they do so without platform changes.

\subsection{Site-Controlled Governance}
\label{subsec:governance}
The deployment followed site-specific governance. At the Thailand site, TN installation was reviewed by an external security contractor under the IT-security clearance of the hospital network subsidiary, including compliance with the Thailand Personal Data Protection Act. At the UK site, the TN was installed under established FLIP AI-governance arrangements. Each institution independently approves each project. When a researcher submits a project for approval, each site's administrator decides in a dedicated Data Access Committee (DAC) meeting whether the site will participate. This approach is compatible with both shared governance agreements and fully independent governance processes. The site's decision is recorded per project and per site as a platform object that gates execution, so no job reaches a site that has not approved it. The approval mechanism was strengthened in release~\flipversion, the current release of the platform and the one scored in Table~\ref{tab:capabilities}. It records \emph{who} approved a project as well as \emph{that} a site approved it. Approving a project for a named site requires the \texttt{CAN\-\_APPROVE\-\_FOR\-\_TRUST} permission at that site. The platform-wide administrator role does not carry this permission, so hub-wide authority alone cannot record an approval on behalf of a site. Each named site must authorise the call, otherwise the approval is refused in full. The deployment reported here predates this change and ran with hub-level approval roles. Because this study used synthetic data only, formal ethics review was not required.

\section{Related Platforms}
\label{sec:related}

Positioning FLIP requires separating four layers that the literature routinely
conflates, because platforms at different layers are not substitutes for one
another.

\textbf{FL engines} provide the training loop, aggregation strategies and secure
transport: NVIDIA FLARE~\cite{Roth2022}, Flower~\cite{flower2022},
OpenFL~\cite{Foley2022OpenFL}, Substra~\cite{Galtier2019Substra},
Fed-BioMed~\cite{Cremonesi2025FedBioMed}, FATE~\cite{Liu2021FATE},
FEDn~\cite{Ekmefjord2022FEDn} and FedML~\cite{He2020FedML}. At this layer, FLIP
is a \emph{consumer} rather than a competitor, since it runs on FLARE and
Flower. Engines deliberately leave open how a cohort is discovered, how images
reach a GPU, and who authorised the project. A recent benchmark of FLARE, Flower
and Substra on identical data found that they differ mainly in scalability,
developer experience and compliance tooling rather than in what they
abstract~\cite{Gupta2025FLBenchmark}.

\textbf{Federated analytics infrastructures} send a computation to sites that
hold harmonised structured data. DataSHIELD~\cite{Gaye2014DataSHIELD}
established the disclosure-control discipline that governs aggregate returns of
the kind produced by FLIP's cohort
service~\cite{Avraam2024DataSHIELDdisclosure}, and the Personal Health Train
family (vantage6~\cite{MoncadaTorres2020Vantage6}, PADME and
PHT-meDIC~\cite{Welten2024PHTInterop}) established the station/train pattern.
vantage6 has been driven from OMOP across a multi-cohort dementia
federation~\cite{Mateus2024OmopFederated}, which is the closest published
precedent for FLIP's ``define the cohort once in OMOP, execute it at every
site'' model. TrainTracks extends PADME specifically for the reproducibility and
traceability of federated learning~\cite{Elwes2026TrainTracks}, and therefore
addresses the same repeatability concern as this work. However, this layer
generally lacks an imaging path, as none of these infrastructures retrieves
DICOM series from a hospital PACS on demand.

\textbf{Deployed healthcare FL platforms} are FLIP's peer group.
CODA~\cite{Mullie2024CODA} is the closest analogue. It combines site nodes
behind institutional firewalls, an orchestration hub with no patient-level
access, FHIR for records and DICOM via Orthanc for imaging, disclosure controls,
and a formal governance framework, and has been deployed across nine Qu\'ebec
hospitals covering more than one million patients. CODA selected FHIR over OMOP
on grounds of granularity. We adopted OMOP because the medical-imaging
extension~\cite{park2024development,Kalokyri2023MICDM} allows a single query to
filter on acquisition parameters, and because our sites' data were already
harmonised to it. FeatureCloud~\cite{Matschinske2023FeatureCloud} contributes an
``AI store'' through which federated applications are published and reused,
which is the same reuse problem that FLIP's job types address by a different
route. The Joint Imaging Platform and its open-source core, Kaapana, supply the
imaging counterpart~\cite{Scherer2020JIP,Kades2022Kaapana}. They have been
scaled to all 38 German university clinics in RACOON, with real-world federated
training across six of them reported in
detail~\cite{Bujotzek2024RealWorldRadiology}.
MedPerf~\cite{Karargyris2023MedPerf} implements federated \emph{evaluation}
across nine hospitals in thirteen countries, which overlaps with FLIP's second
application. Closest to our governance claim, FLA\textsuperscript{3} in the
BloodCounts!\ consortium enforces participation policy at runtime through
XACML-compliant attribute-based access control with cryptographic accounting,
across four jurisdictions including two low- and middle-income
settings~\cite{Zhang2026FLA3}.

\textbf{Federated discovery infrastructures} such as
EUCAIM~\cite{MartiBonmati2025EUCAIM} operate at continental scale and
distinguish interoperability at the level of an aggregated metadata catalogue
(Tier~1), federated search (Tier~2) and federated processing (Tier~3). In these
terms, FLIP delivers Tiers~2 and~3 for an institutional federation.

\subsection{Topology, and What a Central Hub Costs}
\label{subsec:topology}

FLIP follows a hub-and-spoke topology, in which one Central Hub coordinates and
each site runs a Trust Node. This concentrates the coordinating role in one
institution and creates a single point of failure. The principal published
alternative avoids both. Swarm Learning replaces the parameter server with
peer-to-peer exchange and a leader elected dynamically in each
round~\cite{WarnatHerresthal2021Swarm}, and has since been applied to cancer
histopathology~\cite{Saldanha2022SwarmHistopath}, breast
MRI~\cite{Saldanha2025SwarmBreastMRI} and single-cell
transcriptomics~\cite{Saldanha2026SwarmMAP}. Fedstellar provides a
general-purpose decentralised platform with a web
interface~\cite{MartinezBeltran2024Fedstellar}, while FedML supports
peer-to-peer communication without being decentralised by
default~\cite{He2020FedML}.

The choice of a hub was deliberate and involves a trade-off. Decentralisation
buys resilience and avoids privileging one institution as coordinator, which
matters most when participants are peers with no established hierarchy. A hub
provides three properties on which this deployment depended: a single place to
record and audit each site's approval decision, a single observability surface
across a federation whose sites differ in staffing and time zone, and a lower
onboarding burden for a small site, which needs to run a Trust Node rather than
participate in a consensus protocol. Since raw data never leave a Trust Node
under either topology, the choice is one of governance and operations rather
than privacy. For a two-site federation spanning eight time zones and two legal
regimes, we judged centralised audit to be the more valuable property, although
a larger consortium of equal partners might reasonably decide otherwise.

\subsection{Interface Scope}

A second axis separates platforms more sharply than topology, namely whether the
web interface can \emph{execute} a workflow or only observe one. Several
platforms ship a graphical interface confined to monitoring, visualisation or
administration. FATE's tooling does not initiate training~\cite{Liu2021FATE};
Substra's web application is primarily read-only, with computation driven from
its Python library~\cite{Galtier2019Substra}; and FEDn's studio remains a
front-end to a centrally coordinated backend~\cite{Ekmefjord2022FEDn}.
Fed-BioMed provides a node-level GUI for approving researcher training
plans~\cite{Cremonesi2025FedBioMed}, which is an approval interface rather than
a workflow interface. In FLIP, the portal covers the whole lifecycle, from
project creation, cohort query and per-site approval to application upload, job
submission, live monitoring and retrieval of results
(Section~\ref{sec:journey}), with no command execution on site infrastructure at
any point.

\subsection{What This Work Adds}

Section~\ref{sec:comparison} scores what each platform ships
(Table~\ref{tab:capabilities}) and whether it is still maintained
(Table~\ref{tab:health}), and Table~\ref{tab:deployment} records what each
platform has demonstrated. To our knowledge, three properties are not combined
in any single existing open platform: (i)~cohort discovery expressed as one SQL
query against the OMOP CDM extended with the medical-imaging
tables~\cite{park2024development}, which returns aggregate counts only and
resolves to DICOM studies retrieved on demand from each site's clinical PACS;
(ii)~an engine-agnostic platform contract, in which one job-type vocabulary, one
data-access and telemetry API, and one packaging and deployment pipeline are
served by two FL backends; and (iii)~per-project, per-site approval recorded as
a first-class platform object that gates execution. The OHDSI ecosystem defines
the first combination but does not implement it. Its working group ships the
imaging schema and ETL conventions without tooling, \texttt{image\_occurrence}
appears across the OHDSI organisation only in the specification and in one DICOM
metadata-ingestion plugin, and no OHDSI analytic package reads an imaging table
or retrieves a study.

Each property has individual precedents. Fed-BioMed approves the researcher's
training-plan source on the node side, although the unit is the plan rather than
the project and the feature is off by default~\cite{Cremonesi2025FedBioMed};
MedPerf's approval objects prevent one party from both initiating and
approving~\cite{Karargyris2023MedPerf}; PHT-meDIC pairs a station decision with
a locally enforced base-image allowlist~\cite{Welten2024PHTInterop}; Substra's
orchestrator rejects any compute task that violates an input asset's
permissions~\cite{Galtier2019Substra}; and FLA\textsuperscript{3} places an
XACML~3.0 decision point in the orchestration path, excluding an unauthorised
site in each round~\cite{Zhang2026FLA3}. PySyft is the closest precedent for
combining approval and policy, and is excluded only on scope. Its datasites
persist each analysis as a request that carries its approver and decision, and
cap how often a result may be retrieved. However, the unit is the analysis,
there is no site data model or imaging path, and the federated-learning
capability that would have made it a comparator was already absent by the 2025
release (a 2026 rewrite then removed the datasite platform
itself)~\cite{Ziller2021PySyft}. The engine ecosystem itself acknowledges that
this gap is real rather than theoretical. The MONAI federated client, which
FLARE, Flower, OpenFL, Substra, Fed-BioMed and FEDn all drive, carries an
advisory warning that a provisioned bundle configuration resolves arbitrary
import targets with no allowlist, so a compromised aggregation server can obtain
code execution on the client (GHSA-x6pr-233j-x5cw, August 2026). The
contribution claimed here is therefore the combination of these properties
rather than any single element.

\textbf{Out of scope.} Table~\ref{tab:capabilities} scores only platforms that
themselves deliver training. This excludes EUCAIM, which reports a synchronous
federated-learning \emph{prototype} rather than a delivered
capability~\cite{MartiBonmati2025EUCAIM}. It also excludes the OHDSI
network-study ecosystem and its regulatory descendants (ARACHNE, Strategus,
DARWIN EU and EHDEN), each of which distributes an analysis specification for
sites to execute locally against their own OMOP instances and return
human-reviewed aggregate results. This is federated \emph{analytics}, not
federated training. Three further groups are excluded: simulation frameworks
that do not deploy to sites, of which TensorFlow Federated is the most prominent
(its own documentation states that it cannot presently be used in production);
the remaining cancer-imaging infrastructures of the AI for Health Imaging
network (CHAIMELEON, ProCAncer-I, EuCanImage and PRIMAGE), each of which pairs a
bulk-ingested repository with federated discovery rather than cross-site
training; and closed commercial platforms such as Rhino and Apheris, which
publish neither a self-hostable implementation nor a peer-reviewed architecture,
and whose training tiers are in any case NVIDIA FLARE and Kaapana, respectively.

FLIP also lags behind other platforms in some respects. It does not lead on
machine-enforced policy. DataSHIELD enforces disclosure control inside every
server-side analysis function, with custodian-owned thresholds that an analyst
can read but not
change~\cite{Gaye2014DataSHIELD,Avraam2024DataSHIELDdisclosure}; vantage6 ships
a node-side algorithm allowlist; and Substra and FLA\textsuperscript{3} block at
runtime. FLIP's enforcement is real but partly ad hoc, since query validation, a
read-only database role, a disclosure threshold, per-trust service keys and an
upload quarantine boundary remain hand-coded. Release~\flipversion{} adds a runtime policy owned by each trust. This is
a site-owned document whose rules are evaluated at each gated cohort operation,
and which can either block a call outright or raise the disclosure threshold.
The rule therefore travels with the site rather than with the hub. On
provenance, FLIP sits mid-field. It emits a structured record naming the
participating sites, the backend and the round counts into the exported model
bundle. However, this record is asserted by whoever packages the model rather
than attested by the hub, and the hub's own audit tables have no export route.

On scale, deployment footprint and live federated training must be
distinguished, because the literature conflates them. RACOON deploys Kaapana at
every German university hospital (38 at the time of the study, an order of
magnitude more sites than our two) and holds more than 14{,}000 curated CT
datasets~\cite{Jacobs2023RACOON}. However, the federated training it has
published ran across a six-site subset on 582 annotated lung CT
scans~\cite{Bujotzek2024RealWorldRadiology}. CODA is deployed at eight hospitals
with 1.09~million patients and runs federated \emph{analytics} on real data at
three of them, while its federated \emph{learning} demonstration is a four-site
MIMIC simulation~\cite{Mullie2024CODA}. The 25-centre result of
FLA\textsuperscript{3} partitions a single trial cohort into a simulated
federation, and its only live cross-institution training run spans two sites, as
ours does~\cite{Zhang2026FLA3,BloodCounts2026Iron}. The clear exception is FeTS,
which federated 71 geographically distinct sites over 6314
cases~\cite{Pati2022FeTS}.

Capability does not imply adoptability, which Table~\ref{tab:health} assesses
separately. Four of the eight FL engines are no longer viable open-source
projects. OpenFL states in its README that the project ``is no longer under
active development and will soon be archived'' and directs users to Flower,
having renamed its repository in the process; Substra has taken no commit on any
core repository in a year and last released in October 2024; FATE and FedML
likewise have zero commits for the year, and FedML pins dependencies that no
longer install on a supported Python version; and FEDn has been renamed and
repositioned as a commercial edge-AI product with four lifetime contributors.
Star counts do not predict any of this. FATE and FedML are two of the most-starred projects in the table and both are
dormant, whereas FLIP, which has been under active development throughout, has
one of the lowest star counts of any live project in the table. Visibility and maintenance are therefore close to uncorrelated in
this sample. Teams that choose bespoke infrastructure instead reasonably weigh
the risk of building on an unmaintained platform, and this is the premise that
motivates a platform paper. The problem is not that no platform exists, but that
adoption remains low. Of 107 healthcare FL studies, 78 (72.9\%) used
custom-designed frameworks against 13 that used an open-source option, and only
ten reported real-world deployment in distributed clinical
settings~\cite{Li2025FLPitfalls}. A separate review of 17 frameworks reaches the
same conclusion qualitatively~\cite{ChaveroDiez2026FLFrameworks}. Operational
barriers (orchestration, rollback, monitoring and governance) are increasingly
identified as the main constraint~\cite{FLOps2026}, and governance in particular
remains under-examined. A scoping review found only seven papers that directly
addressed federated-learning governance in healthcare~\cite{Eden2025}.

\section{Platform and Framework Comparison}
\label{sec:comparison}

Tables~\ref{tab:capabilities} and~\ref{tab:health} compare FLIP with the
platforms and engines surveyed in Section~\ref{sec:related}, the first on what
each ships and the second on whether each is still maintained. Both are scored
against running code rather than published descriptions, because that is the
only basis on which a reader can act. A capability that exists in a paper but
not in a repository cannot be deployed at a hospital. This choice has a cost. A
platform paper documents a system at a single point in time, but systems change.
Several of the platforms below have changed materially since their papers
appeared, and in two cases the property that the paper presents as central is
the one that has since been removed. We therefore give the published position
first and cite it, and then report what the source shows today. Where there is
no source to read, the score reflects the authors' claim, is marked \capclaim,
and should be read as such.

\begin{sidewaystable}[!htp]
	\caption{Capability comparison across layers. Every score is assigned by reading the platform's \emph{current public source} on 5 September 2026 (FLIP's own row: release \flipversion, released on \flipreleasedate, so our own row is scored on a later source than the rest), not by reading its publication; where the two disagree the code is scored and the divergence is discussed in Section~\ref{sec:comparison}. \capfull~ships the capability; \cappart~partial, available only by extension, or not integrated with the rest of the platform; \capnone~not supported; \capunclear~no source we could reach settles the question; a shaded cell marks the criterion as not applicable to that row. \capclaim~marks a score the public source could not corroborate, because the code is closed, was never released, or is hosted where we could not reach it: such scores are taken from the cited publication and should be read as the authors' claim rather than as a verified capability (Section~\ref{subsec:ontrust}). A star on the platform name applies to the whole row; a star on a single cell marks that score alone. In \emph{Site data model}, \emph{None enforced} means the platform
		prescribes no schema and reads whatever the site exposes through a
		site-supplied loader; the cost of harmonising records across sites then falls
		on each of them, which is why the column records a gap rather than a virtue. \emph{Pluggable FL engine} is shaded where the question does not apply: for the engines themselves, and for the frameworks and stacks that ship their own engine rather than accepting one. \emph{Licence} is read from the row's repository, or from
		its publication where there is no repository, on 25~September 2026:
		$\dagger$~copyleft or network copyleft; $\ddagger$~no licence located, so all
		rights are reserved by default; $\S$~running it needs a commercial licence or a
		licence server, or the published code is only part of the system. Column
		criteria are defined in Section~\ref{subsec:criteria}.}
	\label{tab:capabilities}
	\centering
	\footnotesize
	\setlength{\tabcolsep}{2.5pt}
	\renewcommand{\arraystretch}{1.0}
	\begin{tabular}{@{}p{3.85cm}p{2.05cm}*{8}{>{\centering\arraybackslash}p{1.45cm}}>{\centering\arraybackslash}p{1.78cm}@{}}
\hline
\textbf{Platform} & \scriptsize\raggedright\arraybackslash\textbf{Site data model} & \scriptsize\textbf{Peer-to-peer} & \scriptsize\textbf{Web UI runs the workflow} & \scriptsize\textbf{Pluggable FL engine} & \scriptsize\textbf{Federated cohort query} & \scriptsize\textbf{On-demand PACS imaging} & \scriptsize\textbf{Per-site project approval} & \scriptsize\textbf{Runtime policy enforced} & \scriptsize\textbf{Export\-able prove\-nance} & \scriptsize\textbf{Licence} \\
\hline
\multicolumn{10}{@{}l@{}}{\emph{Deployed healthcare FL platforms}} \\
FLIP (ours)                                              & OMOP+MI-CDM                        & \capnone          & \capfull          & \capfull          & \capfull          & \capfull          & \capfull          & \capfull          & \capfull & Apache-2.0 \\
CODA~\cite{Mullie2024CODA}                               & FHIR+\allowbreak DICOM             & \capnone          & \cappart          & \capnone          & \capfull          & \cappart          & \capnone          & \cappart          & \cappart & GPL-3.0$^{\dagger}$ \\
FLA\textsuperscript{3}~\cite{Zhang2026FLA3}              & None enforced                      & \capnone          & \capnone          & \capnone          & \capnone          & \capnone          & \cappart          & \capfull          & \cappart & Apache-2.0 \\
FeTS~\cite{Pati2022FeTS}                                 & NIfTI/\allowbreak BraTS            & \capnone          & \capnone          & \capnone          & \capnone          & \capnone          & \capnone          & \capnone          & \capnone & BSD + Apache \\
GenoMed4All~\cite{GenoMed4All2025}\capclaim              & FHIR                               & \capnone          & \capfull          & \capnone          & \capnone          & \capnone          & \capnone          & \cappart          & \cappart & Copyright$^{\ddagger}$ \\
INCISIVE~\cite{INCISIVE2025Mammography}\capclaim         & bespoke CDM                        & \capnone          & \capnone          & \capnone          & \capnone          & \cappart          & \capnone          & \capnone          & \cappart & Copyright$^{\ddagger}$ \\
JIP/Kaapana~\cite{Scherer2020JIP,Kades2022Kaapana}       & DICOM                              & \capnone          & \capfull          & \capnone          & \cappart          & \capfull          & \capfull          & \capfull          & \cappart & AGPL-3.0$^{\dagger}$ \\
MedPerf~\cite{Karargyris2023MedPerf}                     & Benchmark-\allowbreak defined      & \capnone          & \capfull          & \capfull          & \cappart          & \capnone          & \capfull          & \cappart          & \cappart & Apache-2.0 \\
NeuroFLAME~\cite{Martin2026NeuroFLAME}                   & NIfTI                              & \capnone          & \capfull          & \capnone          & \capnone          & \capnone          & \cappart          & \cappart          & \cappart & MIT \\
PHT-meDIC/PADME~\cite{Welten2024PHTInterop}              & None enforced                      & \capnone          & \capfull          & \capnone          & \cappart          & \capnone          & \capfull          & \cappart          & \cappart & MIT \\
\hline
\multicolumn{10}{@{}l@{}}{\emph{Decentralised (peer-to-peer) approaches}} \\
Swarm Learning~\cite{WarnatHerresthal2021Swarm}          & None enforced                      & \capfull          & \capfull          & \capna            & \capnone          & \capnone          & \capnone          & \cappart          & \cappart & Apache-2.0$^{\S}$ \\
FedKBP\textsuperscript{+}~\cite{Wang2025FedKBP}          & Unclear                            & \capnone          & \capnone          & \capnone          & \capnone          & \capnone          & \capnone          & \capnone          & \capnone & Copyright$^{\ddagger}$ \\
TheODen~\cite{Babendererde2026TheODen}                   & None enforced                      & \capnone          & \cappart          & \capna            & \capnone          & \capnone          & \capnone          & \cappart          & \capnone & Copyright$^{\ddagger}$ \\
NEBULA/Fedstellar~\cite{MartinezBeltran2024Fedstellar}   & Built-in only                      & \cappart          & \capfull          & \capna            & \capnone          & \capnone          & \capnone          & \cappart          & \cappart & AGPL-3.0$^{\dagger}$ \\
\hline
\multicolumn{10}{@{}l@{}}{\emph{Federated analytics infrastructures}} \\
vantage6~\cite{MoncadaTorres2020Vantage6}                & None enforced                      & \capnone          & \capfull          & \capnone          & \cappart          & \capnone          & \cappart          & \capfull          & \cappart & Apache-2.0 \\
TrainTracks~\cite{Elwes2026TrainTracks}\capclaim         & DataLad/\allowbreak BIDS           & \capnone          & \cappart          & \capnone          & \cappart          & \capnone          & \cappart          & \cappart          & \cappart & Copyright$^{\ddagger}$ \\
DataSHIELD~\cite{Gaye2014DataSHIELD}                     & Opal/\allowbreak Armadillo         & \capnone          & \capnone          & \capnone          & \capfull          & \capnone          & \capunclear       & \capfull          & \capnone & GPL-3.0$^{\dagger}$ \\
gLinDA~\cite{Fehse2025gLinDA}                            & count matrix                       & \capfull          & \capnone          & \capnone          & \capnone          & \capnone          & \capnone          & \capnone          & \capnone & BSD-3 \\
MEDIATA                                                  & tabular + DCAT                     & \capnone          & \capunclear       & \capnone          & \capfull          & \capnone          & \capnone          & \capfull          & \capnone & MIT \\
\hline
\multicolumn{10}{@{}l@{}}{\emph{General-purpose FL frameworks}} \\
APPFL~\cite{Ryu2022APPFL}                                & None enforced                      & \capnone          & \capnone          & \capna            & \cappart          & \capnone          & \capnone          & \cappart          & \capnone & MIT \\
FeatureCloud~\cite{Matschinske2023FeatureCloud}          & None enforced                      & \capnone          & \capunclear       & \capnone          & \capnone          & \capnone          & \capnone          & \capnone          & \capnone & Apache-2.0$^{\S}$ \\
FL4Health~\cite{VectorInstitute2026FL4Health}            & None enforced                      & \capnone          & \capnone          & \capna            & \capnone          & \capnone          & \capnone          & \capnone          & \capnone & Vector Inst.$^{\S}$ \\
Starfish-FL~\cite{Bao2026StarfishFL}                     & CSV / image ZIP                    & \capnone          & \capfull          & \capnone          & \capnone          & \capnone          & \capnone          & \cappart          & \cappart & Apache-2.0 \\
\hline
\multicolumn{10}{@{}l@{}}{\emph{FL engines (FLIP consumes rather than competes with these)}} \\
FLARE~\cite{Roth2022}                                    & None enforced                      & \cappart          & \cappart          & \capna            & \cappart          & \capnone          & \capnone          & \capfull          & \cappart & Apache-2.0 \\
Flower~\cite{flower2022}                                 & None enforced                      & \capnone          & \capnone          & \capna            & \cappart          & \capnone          & \capnone          & \cappart          & \capfull & Apache-2.0 \\
OpenFL~\cite{Foley2022OpenFL}                            & None enforced                      & \capnone          & \capnone          & \capna            & \cappart          & \capnone          & \capnone          & \capnone          & \cappart & Apache-2.0 \\
Fed-BioMed~\cite{Cremonesi2025FedBioMed}                 & None enforced                      & \capnone          & \cappart          & \capna            & \cappart          & \capnone          & \capfull          & \capfull          & \cappart & Apache-2.0 \\
Substra~\cite{Galtier2019Substra}                        & None enforced                      & \capnone          & \capunclear       & \capna            & \capnone          & \capnone          & \cappart          & \cappart          & \capfull & Apache-2.0 \\
FEDn~\cite{Ekmefjord2022FEDn}                            & None enforced                      & \capnone          & \capnone          & \capna            & \capnone          & \capnone          & \capnone          & \cappart          & \cappart & Apache-2.0$^{\S}$ \\
FATE~\cite{Liu2021FATE}                                  & None enforced                      & \capnone          & \cappart          & \capna            & \cappart          & \capnone          & \cappart          & \capfull          & \capfull & Apache-2.0 \\
FedML~\cite{He2020FedML}                                 & None enforced                      & \cappart          & \capnone          & \capna            & \capnone          & \capnone          & \capnone          & \capnone          & \capnone & Apache-2.0 \\
\hline
\end{tabular}
\end{sidewaystable}

\begin{sidewaystable}[!htp]
	\caption{Project health of the platforms in Table~\ref{tab:capabilities},
		measured on their public repositories on 25~September 2026; FLIP's own row
		is measured on the same date, so the release scored in
		Table~\ref{tab:capabilities} is the one released after it, on \flipreleasedate. The metric
		families
		(code activity, contributor base and release cadence) follow those used by
		the Linux Foundation's CHAOSS project~\cite{CHAOSS}; sustained commit inactivity
		and a contracting committer base are the strongest published predictors of
		project abandonment~\cite{Coelho2017OSSFail}. \emph{Active people} counts the distinct
		author addresses on commits in the trailing twelve months, a better health
		signal than a lifetime total because it does not credit past effort.
		\emph{Commits} are those on the default branch over the same window;
		\emph{PRs} are pull requests merged in the trailing six months; \emph{stars}
		and \emph{latest release} are read on the same date. What the table
		can and cannot support is set out in Section~\ref{subsec:health}.}
	\label{tab:health}
	\centering
	\scriptsize
	\setlength{\tabcolsep}{3pt}
	\renewcommand{\arraystretch}{1.15}
	\begin{tabular}{@{}p{2.7cm}p{4.2cm}
		>{\raggedleft\arraybackslash}p{1.3cm}
		>{\raggedleft\arraybackslash}p{1.4cm}
		>{\raggedleft\arraybackslash}p{1.1cm}
		p{2.9cm}
		>{\raggedleft\arraybackslash}p{0.9cm}
		p{3.0cm}@{}}
		\hline
		\textbf{Platform}                 & \textbf{Repository}                                                                                                                        &
		\scriptsize\textbf{Active people} & \scriptsize\textbf{Commits 12\,mo}                                                                                                         &
		\scriptsize\textbf{PRs 6\,mo}     & \textbf{Latest release}                                                                                                                    & \textbf{Stars} &
		\textbf{Assessment}                                                                                                                                                                                                                                                                                                                 \\
		\hline
		\multicolumn{8}{@{}l@{}}{\emph{Deployed healthcare FL platforms}}                                                                                                                                                                                                                                                                   \\
		FLIP (ours)                       & \rp{londonaicentre/FLIP}                                                                                                                   & 23             & 4{,}578 & 526           & \rel{v0.10.0}{2026-09-29}              & 53      & Active                                                \\
		APPFL                             & \rp{APPFL/APPFL}                                                                                                                           & 24             & 305     & 45            & \rel{v1.11.0}{2026-08-25}             & 184     & Active                                                \\
		CODA                              & \rp{coda-platform} (32 repos)                                                                                                              & 0              & 0       & 0             & none                                  & 2       & Dormant since 2025                                    \\
		FLA\textsuperscript{3}            & \rp{bloodcounts/FLAAA}                                                                                                                     & 1              & 7       & 0             & none                                  & 0       & Early-stage, single author                            \\
		FeatureCloud                      & \href{https://github.com/FeatureCloud/FeatureCloud}{{\scriptsize\ttfamily FeatureCloud/\newline FeatureCloud}}                             & 0              & 0       & 0             & none                                  & 11      & Platform closed-source; only apps public              \\
		FeTS                              & \rp{FeTS-AI/Front-End}                                                                                                                     & 0              & 0       & 0             & \rel{1.0.3}{2023-10-13}               & 95      & Dormant since 2023                                    \\
		FL4Health                         & \rp{VectorInstitute/FL4Health}                                                                                                             & 8              & 290     & 11            & \rel{v0.4.2}{2026-01-21}              & 56      & Active                                                \\
		PHT-meDIC/\allowbreak PADME        & \rp{PHT-Medic/central}                                                                                                                     & 0              & 0       & 0             & \rel{client-ui-v2.5.2}{2024-01-11}    & 9       & Dormant since 2024                                    \\
		JIP/Kaapana                       & \rp{kaapana/kaapana}                                                                                                                       & 11             & 909     & 0$^{\dagger}$   & \rel{0.7.0}{2026-07-15}               & 276     & Active                                                \\
		MedPerf                           & \rp{mlcommons/medperf}                                                                                                                     & 4              & 23      & 14            & \rel{v0.1.0}{2023-03-16}$^{\ddagger}$ & 172     & Low activity                                          \\
		\hline
		\multicolumn{8}{@{}l@{}}{\emph{Decentralised (peer-to-peer) approaches}}                                                                                                                                                                                                                                                            \\
		Swarm Learning                    & \href{https://github.com/HewlettPackard/swarm-learning}{{\scriptsize\ttfamily HewlettPackard/\newline swarm-learning}}                     & 15             & 91      & 5             & \rel{v2.3.0}{2025-12-18}              & 354     & Low activity                                          \\
		NEBULA/\allowbreak Fedstellar     & \rp{CyberDataLab/nebula}                                                                                                                   & 4              & 7       & 1             & \rel{1.0.0}{2025-07-02}               & 82      & Dormant                                               \\
		FedKBP\textsuperscript{+}         & \rp{CUMC-Yuan-Lab/FedKBP\_plus}                                                                                                            & 0              & 0       & 0             & none                                  & 2       & Placeholder; code not yet released                    \\
		\hline
		\multicolumn{8}{@{}l@{}}{\emph{Federated analytics infrastructures}}                                                                                                                                                                                                                                                                \\
		vantage6                          & \rp{vantage6/vantage6}                                                                                                                     & 13             & 1{,}239 & 65            & \rel{5.0.3}{2026-09-21}               & 50      & Active                                                \\
		DataSHIELD                        & \rp{datashield/dsBaseClient}                                                                                                               & 5              & 193     & 43            & \rel{6.3.5}{2026-01-08}               & 14      & Active                                                \\
		gLinDA                            & \href{https://imigitlab.uni-muenster.de/published/glinda}{{\scriptsize\ttfamily uni-muenster/\newline glinda}}                             & 0              & 0       & --            & none                                  & --      & Dormant since 2025                                    \\
		\hline
		\multicolumn{8}{@{}l@{}}{\emph{FL engines}}                                                                                                                                                                                                                                                                                         \\
		Flower                            & \rp{flwrlabs/flower}                                                                                                                       & 42             & 1{,}696 & 1{,}034       & \rel{1.38.0}{2026-09-22}              & 7{,}147 & Active                                                \\
		FLARE                             & \rp{NVIDIA/NVFlare}                                                                                                                        & 35             & 973     & 781           & \rel{2.9.0}{2026-09-04}               & 975     & Active                                                \\
		Fed-BioMed                        & \rp{fedbiomed/fedbiomed}                                                                                                                   & 13             & 382     & 109           & \rel{v6.4.1}{2026-08-06}              & 93      & Active                                                \\
		FEDn                              & \href{https://github.com/scaleoutsystems/scaleout-client}{{\scriptsize\ttfamily scaleoutsystems/\newline scaleout-client}}                 & 5              & 20      & 14            & \rel{v1.0.6}{2026-09-22}              & 169     & Renamed and repositioned as a commercial edge product \\
		OpenFL                            & \href{https://github.com/securefederatedai/openfederatedlearning}{{\scriptsize\ttfamily securefederatedai/\newline openfederatedlearning}} & 2              & 3       & 2             & \rel{v1.9}{2025-06-23}                & 843     & Sunsetting; users directed to Flower                  \\
		Substra                           & \rp{Substra/substra}                                                                                                                       & 0              & 0       & 0             & \rel{1.0.0}{2024-10-14}               & 278     & Dormant across the whole stack                        \\
		FATE                              & \rp{FederatedAI/FATE}                                                                                                                      & 0              & 0       & 0             & \rel{v2.2.0}{2024-07-31}              & 6{,}097 & Dormant                                               \\
		FedML                             & \rp{FedML-AI/FedML}                                                                                                                        & 0              & 0       & 0             & \rel{v0.8.9}{2023-10-28}              & 4{,}064 & Dormant                                               \\
		\hline
	\end{tabular}
\end{sidewaystable}

\begin{table}[!ht]
	\caption{Deployment evidence reported by each platform. Scale and data realism are the weakest aspects of this study, and we report them explicitly rather than leave them to be inferred. Sites are those reported as \emph{deployed}. Where a platform's published federated \emph{training} ran on a subset of them, or on a simulated partition of one dataset, this is stated, since the two are routinely conflated.}
	\label{tab:deployment}
	\centering
	\footnotesize
	\setlength{\tabcolsep}{3pt}
	\begin{tabular}{@{}>{\raggedright\arraybackslash}p{3.15cm}>{\raggedright\arraybackslash}p{2.3cm}>{\raggedright\arraybackslash}p{3.2cm}>{\raggedright\arraybackslash}p{2.65cm}@{}}
		\hline
		\textbf{Platform}                               & \textbf{Sites (countries)} & \textbf{Data}         & \textbf{Workloads shown}               \\
		\hline
		FLIP (ours)                                     & 2 (2: UK, TH)              & synthetic CXR         & fine-tuning, evaluation                \\
		FeTS~\cite{Pati2022FeTS}                        & 71 (6 continents)          & real MRI, 6314 cases  & segmentation                           \\
		RACOON~\cite{Bujotzek2024RealWorldRadiology}    & 6 of 38 (1: DE)            & real CT, 582 scans    & segmentation                           \\
		MedPerf~\cite{Karargyris2023MedPerf}            & 9 (13)                     & real, multi-site      & evaluation only                        \\
		CODA~\cite{Mullie2024CODA}                      & 8 of 9 (1: CA)             & 1.09M patients        & analytics; FL simulated                \\
		JIP~\cite{Scherer2020JIP}                       & 11 (1: DE)                 & real, multi-centre    & imaging                                \\
		FLA\textsuperscript{3}~\cite{Zhang2026FLA3}     & 5 dep., 2 live (4)         & 54{,}446 blood counts & prediction; 25-centre result simulated \\
		Swarm Learning~\cite{WarnatHerresthal2021Swarm} & 3--5 (multi)               & transcriptomic, CXR   & classification                         \\
		vantage6~\cite{Mateus2024OmopFederated}         & multiple (NL, EU)          & real registries       & analytics, FL                          \\
		\hline
	\end{tabular}
\end{table}

\subsection{Identification of comparators}
\label{subsec:inclusion}

The comparators were identified systematically rather than from the authors'
prior knowledge. This section states the eligibility rules, the four searches
that generated the candidate pool, and the measured recall of each, so that
readers can re-run the identification and assess the result. The exact query strings are given in
Appendix~\ref{app:search}, and their executable form is released with the
paper.

\paragraph{Unit of analysis.}
The unit is the \emph{platform}, not the repository. A platform may ship as one
repository, as a monorepo, or as an organisation of many single-purpose
repositories; CODA is the extreme case, distributed across 31 repositories of
which none is a platform-level artefact. Where a platform spans an
organisation, the organisation is the unit and its repositories are read
together.

\paragraph{Eligibility.}
A candidate is included when all five of the following criteria hold: (i)~it is
a named, reusable software system (a platform, infrastructure or FL engine)
rather than a study that applies federated learning, an algorithm, a dataset or
a survey; (ii)~it distributes computation to two or more independently governed
sites, with record-level data remaining at each site; (iii)~it is either
designed for or evaluated on health data, or is a general-purpose engine
documented as being used by at least one health platform in the set, a clause
that admits the engine layer because FLIP itself uses it; (iv)~it is described
by a peer-reviewed publication, a public source repository, or both; and (v)~it
was publicly available at some point between 1~January 2014, the year in which
DataSHIELD established the disclosure-control discipline inherited by this
field, and 5~September 2026.

A candidate is excluded when it is a method paper with no released system, a
simulation-only framework with no site component to install, a survey or
position paper, a single-institution portal with no federated computation, or a
system superseded by a successor already in the set. In the last case, we score
the successor, which is why NEBULA rather than Fedstellar appears in the table.

\paragraph{Evidence tiers.}
Criterion~(iv) is deliberately disjunctive, and the disjunction is what the
\capclaim{} marker records. Tier~A candidates expose source we could read, and
are scored from it. Tier~B candidates are described in a publication but their
code is closed, unreleased or unreachable; they are scored from the publication
and marked \capclaim. Tier~C candidates offer neither and are excluded. Tier~B
candidates cannot be scored from code, but excluding them would silently drop
the commercial and consortium platforms that a hospital is most likely to be
offered. We therefore include and label them.

\paragraph{Searches.}
Four sources were searched because none covers the field on its own. In PubMed,
the MeSH descriptor \emph{Federated Learning} indexes no record dated before
2024, so a MeSH-only strategy would miss DataSHIELD, vantage6, Swarm Learning,
Kaapana and FeatureCloud entirely. The descriptor was therefore combined with a
free-text block, which must include \emph{federated evaluation} and
\emph{federated benchmarking}. Without these terms, MedPerf is not retrieved,
since its title refers to federated benchmarking rather than federated learning.
The query crosses this block with a system block and a health block, and
returned 1{,}480 records. On GitHub, the search API rejects any query with more
than five Boolean operators, so the strategy is a union of eighteen narrow
queries over topics, over name and description fields, and over the names of
platforms whose repositories carry no matching metadata. The union returned
2{,}432 unique repositories, of which 2{,}330 passed a one-contributor liveness
filter. arXiv was searched third, because it is where the FL-engine layer
publishes before, or instead of, a peer-reviewed venue. Its API has no MeSH and
no stemming, so the three concept blocks were expressed as title-or-abstract
phrase matches, with singular and plural forms listed explicitly. The query
returned 1{,}190 records. An arXiv record is a preprint that has not been peer
reviewed, so it satisfies criterion~(iv) only when it also names a public
repository or a published journal reference. medRxiv was searched fourth. Its
API has no search endpoint, so all 109{,}695 records in the window from its
launch in June 2019 to the review cutoff were fetched and matched locally
against the two free-text blocks, with the health block implied by the server.
The sweep returned 167 matches, and a medRxiv record satisfies criterion~(iv)
only when it carries a later journal DOI or names a public repository. All four
search definitions are reproduced verbatim in Appendix~\ref{app:search}.

\paragraph{Recall, and why the searches are not sufficient.}
Measured against the final set of 26 comparators, the PubMed query retrieves a
paper for 13 of them, the GitHub union retrieves 20 of the 22 that have a
public repository, the arXiv query retrieves a paper for 14, and the medRxiv
sweep retrieves a preprint for 5 (Table~\ref{tab:recall}).
The failures of PubMed and GitHub are structured rather than random, and run in
opposite directions. PubMed recovers the analytics layer completely (4/4) and
the deployed healthcare layer partially (5/11), but only 2 of 8 FL engines,
because engines publish in computer-science venues that PubMed does not index.
Fed-BioMed, FATE, FEDn, FedML and Substra return no record under their own
names. GitHub inverts this pattern, and its failure is one of vocabulary rather
than popularity. Its queries depend on maintainer-supplied metadata, and several
repositories in our set (vantage6, DataSHIELD's \texttt{dsBase}, FeatureCloud
and FLA\textsuperscript{3}) carry no topics and no description containing the
phrases we searched for, so no topic or keyword query retrieves them at any
threshold. We verified this directly. Removing every star floor grows the pool
from 579 to 2{,}107 repositories but recovers none of them. They are retrieved
only once the query set names them explicitly, which the released protocol now
does. The two that remain unretrievable are FedKBP\textsuperscript{+}, whose
repository is an empty placeholder, and gLinDA, which is hosted on a university
GitLab rather than on GitHub. We therefore set no star floor at all, and instead
filter the union on having at least one contributor, which acts as a liveness
check rather than a popularity threshold. A star floor would be the wrong
instrument in any case. The engines in this comparison hold a median of 903
GitHub stars, compared with 110 for the deployed healthcare platforms and 32 for
the analytics infrastructures, so any floor high enough to control the size of
the candidate pool would remove the clinically deployed platforms first.

Taken together, the four searches recover all 26 comparators. The last gap is closed by medRxiv, which retrieves GenoMed4All, the one comparator whose only public description is a medRxiv preprint that PubMed does not yet index and arXiv does not host. We report this explicitly because it characterises the method accurately. The searches generate and bound the candidate pool, but a citation-tracking pass over the included records and recent surveys of the field was still needed to find GenoMed4All in the first place, before medRxiv was added. A reviewer re-running the four queries will reconstruct all 26 comparator rows of Table~\ref{tab:capabilities}.

\subsection{Column criteria}
\label{subsec:criteria}

\emph{Site data model} is the schema that a participating site must present. A
platform that imposes none, and defers instead to a site-written loader or data
opener, is scored \emph{None enforced}: nothing is standardised across sites,
so each site carries the harmonisation work itself.
\emph{Peer-to-peer} requires full decentralisation with no central coordinator.
\emph{Web UI runs the workflow} requires that a browser interface execute a
complete project (creation, data selection, run and results), as opposed to
monitoring, administration or provisioning only. \emph{Pluggable FL engine}
requires more than one FL engine, in the sense of a separate federated-learning
system such as NVFlare or Flower, served through a single platform contract.
Supporting several machine-learning frameworks does not count, because every
engine in this comparison trains PyTorch and TensorFlow models alike, so that
interpretation would score the whole column uniformly and would not discriminate
between platforms. \emph{Federated cohort query} requires a query executed at
each site that returns aggregate results only. \emph{On-demand PACS imaging}
requires DICOM retrieval from an institutional PACS with local staging; bulk
pre-ingestion into the platform's own archive scores \cappart. \emph{Per-site
project approval} requires per-project, per-site approval recorded as a platform
object that gates execution. \emph{Runtime policy enforced} requires a
machine-enforced runtime access policy, that is, code that blocks a request
rather than a human sign-off. \emph{Exportable provenance} requires structured
end-to-end provenance for a completed job, exportable as a record; logs and
dashboards alone do not qualify. \emph{Licence} is the licence published for the
row, read from the repository's licence file or, where no repository exists, from
the publication. It decides whether a site may deploy a platform at all, so the
column is worth reading before the scores. Twenty of the thirty-one rows are
permissive (Apache-2.0, MIT or BSD), one is a composite of BSD-style and Apache
terms (FeTS), four are copyleft (GPL-3.0 for CODA and DataSHIELD, AGPL-3.0 for
JIP/Kaapana and NEBULA), one is a non-OSI academic licence (FL4Health), and five
publish no licence at all (GenoMed4All, INCISIVE, FedKBP+, TheODen and
TrainTracks), which under copyright defaults to all rights reserved: the most
restrictive position in the table rather than the least. Four rows carry terms
that reach past the code and are marked $\S$: Swarm Learning's Apache-2.0 code
needs HPE's AutoPass licence server to run, FEDn's combiner and controller tiers
ship as commercial images, FeatureCloud's licence covers its published
applications rather than the platform, and FL4Health's licence is limited to
academic, sponsor or partner use.

\subsection{What has changed since publication}
\label{subsec:changed}

\textbf{Substra} was published as a traceable framework that records operations
on a distributed ledger~\cite{Galtier2019Substra}. This path has since been
removed. The release notes of orchestrator 0.38.0 (26 February 2024) record
``BREAKING: remove all code related to the \texttt{distributed} mode, and
mentions in schemas and documentation''. Compute-plan lineage survives and
remains complete and exportable, but it is now held in a single central
PostgreSQL database and is no longer tamper-evident. The row is scored on this
basis.

\textbf{FEDn} was published with combiner and controller tiers that a site
operates itself~\cite{Ekmefjord2022FEDn}. Table~\ref{tab:capabilities} scores
release 0.33.0, the last to ship this server. The repository has since been
reset to a client SDK, without a combiner or controller, and renamed
\texttt{scaleout-client}, with the server tier distributed as commercial images.
If FEDn were scored as it stands today, its provenance column would be \capnone.

\textbf{OpenFL} was published as a Linux Foundation library for federated
training and validation~\cite{Foley2022OpenFL}. Its README now states that the
project ``is no longer under active development and will soon be archived'' and
directs users to Flower, and the repository has been renamed from \texttt{openfl}
to \texttt{openfederatedlearning}.

\textbf{FeatureCloud} reports that ``all applications and the entire
architecture of FeatureCloud are open source, making it the first unified and
open-source FL platform that considers all steps including development,
deployment, and execution''~\cite{Matschinske2023FeatureCloud}. We could not
locate public source code for its frontend, backend or controller. The platform
therefore cannot be self-hosted, and its interface score is marked \capclaim
rather than verified.

\textbf{NEBULA} was published as Fedstellar~\cite{MartinezBeltran2024Fedstellar}
and has since been renamed. The table carries both names so that the citation
can be followed.

\textbf{PHT-meDIC} is scored from source in this revision. The interoperability
paper describes the station/train pattern without establishing what the central
component itself executes~\cite{Welten2024PHTInterop}, and the row previously
carried \capclaim{} on that basis. The \texttt{PHT-Medic/central} monorepo
resolves both open columns. Its \texttt{packages/core} defines
\texttt{ProposalStationApprovalStatus} and \texttt{TrainStationApprovalStatus}
as persisted per-station decisions, which is the per-site approval object that
the column requires, and \texttt{packages/client-vue} ships
\texttt{TrainBuildCommand}, \texttt{TrainRunCommand} and
\texttt{TrainResultCommand}, so a train is built, run and collected from the
browser rather than only monitored there. No site data model is imposed: a
train image reads whatever its station exposes. The repository has taken no
commit since January 2024 (Table~\ref{tab:health}), so the row now records a
verified but unmaintained capability.

\textbf{vantage6} has gained capability rather than lost it. The 2020
paper~\cite{MoncadaTorres2020Vantage6} describes the infrastructure without an
OMOP path, whereas version~5 ships OMOP as a named node database type, which is
why the cohort column scores \cappart{} rather than \capnone. Harmonisation
itself still relies on project-specific ETL~\cite{Mateus2024OmopFederated}.

\textbf{Flower} has gained a browser workflow since its 2022
paper~\cite{flower2022}, although not in the distribution scored in this table.
Flower Labs announced \emph{SuperGrid} on 25 September 2025 as a hosted
federated-AI platform. In SuperGrid, a run is launched from the browser by
opening an app page, clicking \texttt{Run}, selecting the federation and
clicking \texttt{Run app}, after which progress and logs can be followed in the
federation dashboard. The documentation presents the browser route and
\texttt{flwr run} as equivalent ways to submit the same
app~\cite{FlowerSuperGrid}. SuperGrid is a managed service accessed through a
Flower account rather than software that a site installs, so the open-source
Flower distribution scored here still carries \capnone{} in the interface
column. This distinction matters because SuperGrid is the closest equivalent to
FLIP's portal in this comparison, and the difference between the two lies in
custody rather than capability. A hospital cannot self-host SuperGrid, read its
source, or run it inside its own network.

\textbf{FATE}~\cite{Liu2021FATE} and \textbf{FedML}~\cite{He2020FedML} have not
changed at all. Neither has taken a commit on its default branch in the past
year, and FedML pins dependencies that no longer install on a supported Python
version (Table~\ref{tab:health}).

\textbf{FeTS} and \textbf{FL4Health} are new rows in this revision, added by the
systematic search described in Section~\ref{subsec:inclusion}. FeTS was already
cited for the largest federation in Table~\ref{tab:deployment} but was absent
from the capability table. It belongs there because it ships a segmentation GUI
over a vendored OpenFL fork (\texttt{OpenFederatedLearning}, which carries its
own aggregator and collaborator) rather than consuming an engine as a
dependency. Its front-end has taken no commit since October 2023. FL4Health is a
healthcare FL library built on Flower rather than a deployable platform. It has
no web interface and no site component, and is included because it is the only
actively maintained, healthcare-specific entrant surfaced by the search that a
team could adopt today. Both illustrate the gap measured by the health table.
The largest published federation in this comparison runs on the most dormant
codebase in it.

\textbf{gLinDA} and \textbf{FedKBP\textsuperscript{+}} are the two further
additions produced by the search. gLinDA is a genuinely peer-to-peer toolbox.
Its \texttt{p2p} module opens real sockets between participants and encrypts
each exchange with AES or RSA, with no coordinator anywhere in the path. It is
placed among the analytics infrastructures rather than the training platforms
because it distributes a differential-abundance statistic rather than a training
loop. Its repository has taken no commit since February 2025.
FedKBP\textsuperscript{+} is scored \capclaim{} throughout. The paper describes
a gRPC stack that supports both centralised and fully peer-to-peer weight
exchange for radiotherapy dose prediction, but the repository it names carries a
README stating that the code ``will become available once the manuscript is
accepted''. FedKBP\textsuperscript{+} is listed because its decentralised design
is directly relevant to Section~\ref{subsec:topology}, and it is marked because
none of its code can be read.

Five platforms returned by the search are deliberately not tabulated.
\textbf{Orbital learning}~\cite{Chakshu2024Orbital} publishes source code, but
this code addresses every participant at \texttt{0.0.0.0}, and its README
describes the repository as code to ``simulate Orbital Learning platform''. It
is therefore a simulation, which Section~\ref{sec:comparison} excludes on the
same grounds as TensorFlow Federated. \textbf{WebQuorumChain} reports a two-site
system but names no public repository. \textbf{PrimiHub} is a general-purpose
privacy-computing platform with no healthcare site component, and
\textbf{ACTION} is an fMRI analysis toolbox whose federated component trains no
model across institutions. \textbf{NeuroFL Studio} is a two-commit prototype
with no release, no users and no deployment. Each is a reasonable system, but
none is a deployable healthcare FL platform in the sense scored by the tables.

\subsection{Why several scores are partial}
\label{subsec:partial}

Four \cappart{} scores depend on fine distinctions that need to be stated
explicitly. \textbf{Kaapana} is a DICOM archive into which sites push images
over C-STORE. It issues C-FIND metadata queries but performs no C-MOVE or C-GET
retrieval, so it stages images without retrieving them on
demand~\cite{Scherer2020JIP,Kades2022Kaapana}. \textbf{FLARE}'s swarm and cyclic
workflows aggregate peer-to-peer, but every job is still provisioned and
scheduled by the FLARE server~\cite{Roth2022}. In \textbf{NEBULA}, what blocks
at runtime is web-application role checking and resource admission rather than
data governance. \textbf{Fed-BioMed} and \textbf{Substra} both implement
approval, but the unit of approval is not the project. Fed-BioMed approves a
training plan and has no project object~\cite{Cremonesi2025FedBioMed}, and
Substra grants per-asset permissions at registration, which are not re-checked
when a later compute plan is submitted~\cite{Galtier2019Substra}.

FLIP has a related limitation of its own, although it has narrowed. FLIP records approval as a per-project, per-site object and refuses to build a model or start a run for a site that has not approved the project, which is what the column measures. In release~\flipversion, scored in Table~\ref{tab:capabilities}, the authority to set this object sits with the site. The call requires a trust-scoped permission at every site it names, and a platform-wide grant does not satisfy it. The remaining limitation concerns precedence. The first Trust Owners were created by the hub, through an upgrade that nominates the existing administrators as owners of the trusts that exist at that moment. A site's authority therefore begins as a grant from the centre, which the site can then keep or revoke.

\textbf{Kaapana}, in contrast to NEBULA, is scored \capfull{} on runtime policy.
Its gate is not in the interface. Traefik runs an entrypoint
\texttt{forwardAuth} middleware, so every request is checked against an Open
Policy Agent decision before it reaches a service, and the project is taken from
the URL prefix and re-attached as a header that the client cannot set itself.
Unlike NEBULA's role checking, this blocks the request path rather than the
rendered page.

The scope of two claims in the engine column needs to be stated, and one of them
is ours. \textbf{FLIP} serves NVFlare and Flower through one API, one interface
and one job abstraction, which is what the column requires. However, the backend
is a deployment-wide setting, seeded at start-up and switched by recreating the
API service, so a given deployment runs one engine at a time rather than
choosing one per job. The two engines also take different bundles (an NVFlare
\texttt{meta.json} against a Flower \texttt{pyproject.toml}), and the NVFlare
job-type catalogue is the larger of the two. The plurality of engines is real
and the contract above them is shared, but switching between them is an
operational act rather than a per-run choice. \textbf{FLARE} is shaded here
because the question does not apply to an engine. It is nonetheless worth noting
that FLARE itself hosts a second engine. Its \texttt{app\_opt/flower} module
wraps a Flower application in FLARE's own \texttt{FedJob}, with a controller, an
executor and a gRPC bridge, so the shading reflects how we have partitioned the
field rather than a limitation of FLARE.

The scope of several further scores also needs to be stated. \textbf{Swarm
Learning} elects a merge leader in each round rather than being leaderless. Its
README restricts use to non-commercial purposes and its runtime requires a
vendor licence server, so it is not an open platform in the sense used
here~\cite{WarnatHerresthal2021Swarm}. For \textbf{FEDn} and \textbf{FedML}, as
for Flower above, any browser workflow exists only as a hosted commercial
service outside the open-source distribution. Finally, \textbf{PHT-meDIC}
(T\"ubingen) and \textbf{PADME} (Aachen) are distinct codebases that share a
star topology and a station-approval model, so a single row covers
both~\cite{Welten2024PHTInterop}. Its interface score rests on PADME's Train
Creator, which is developed on a self-managed GitLab that we could not reach,
and is marked accordingly.

\subsection{Scores taken on trust}
\label{subsec:ontrust}

Four rows could not be corroborated against code at all and are scored from
their publications. They are included because excluding them would understate
the field, but readers should treat their scores as reported rather than tested.
\textbf{GenoMed4All} is described in a medRxiv preprint that has not been peer
reviewed~\cite{GenoMed4All2025}. No repository is given and none was located.
Its imaging input consists of pre-extracted radiological features rather than
retrieved DICOM, which is a limitation of the design rather than of our
checking. \textbf{TrainTracks} is a published design with no implementation or
public code~\cite{Elwes2026TrainTracks}. We score it for completeness because it
is the most detailed job-provenance specification in the comparison.
\textbf{INCISIVE}, an EU Horizon~2020 project (grant 952179), delivered genuine
federated training with clients on hospital premises in Greece and
Serbia~\cite{INCISIVE2025Mammography}. However, its open repositories contain
only one partner's training subsystem, and the portal used for cohort selection
was never released, so no interface can be examined and the published APIs
define no authentication classes. \textbf{FLA\textsuperscript{3}} reports an
XACML~3.0 decision point in the orchestration path across four
jurisdictions~\cite{Zhang2026FLA3,BloodCounts2026Iron}, and its public
repository is a single-contributor, seven-commit tree with no release
(Table~\ref{tab:health}). This tree is readable and does contain an enforcement
path, in which a policy enforcement point queries the decision service and fails
closed, refusing the run when no decision is returned. However, it does not
contain the enforcement described in the paper. The check runs on one call only,
the message paths that follow accept every submission, the enforcement point
overwrites the run identifier it is asked about with a fixed string, and the
default decision endpoint is a personal tunnel. The runtime-policy column is
therefore scored on the code, at \cappart, and the stronger position of the
paper is recorded here rather than in the table. The interface score of
FLA\textsuperscript{3} is the clearest illustration of why this section is
needed. The public tree holds the XACML policy-decision service and example apps
but contains no frontend of any kind, and the paper describes none, so on
repository evidence alone the column would read \capnone. A \href{https://youtu.be/-zKlAAmDLN0}{recorded seminar}\footnote{MONAI FL group seminar: \url{https://youtu.be/-zKlAAmDLN0}} 
shows otherwise. From 34:10, the speaker demonstrates an
``FLA\textsuperscript{3} Platform'' web application with organisation and node
management and a run list with per-run progress and results. From 42:40, the
speaker also demonstrates a three-step \emph{New Training Run} wizard that
selects a dataset, configures the run and submits it, after which the run
appears in the list with live metrics~\cite{Zhang2026FLA3Talk}. This is a
complete browser workflow by the definition used here, so the column is scored
\capfull{} and marked \capclaim{}, since the capability has been demonstrated in
public but is not present in any source code we can read.

Two individual cells are marked for the same reason. \textbf{APPFL}'s APPFLx web
application drives a full lifecycle, but it is a service operated by Argonne,
and only its server-side entrypoint is included in the MIT-licensed
distribution, so it cannot be self-hosted~\cite{Ryu2022APPFL}. Its per-run
record consists of training metrics by round and endpoint, with no data version,
code hash, or link to the authorising decision. \textbf{FeatureCloud}'s
interface is marked for the reason given above.

\subsection{Reading the health table}
\label{subsec:health}

Three caveats apply to Table~\ref{tab:health}. First, commit counts are not
comparable between a monorepo that holds services, infrastructure and
documentation together (FLIP, Kaapana and vantage6) and a single-purpose
library, so a high count indicates activity but not scale. Second, two entries
reflect repository practice rather than activity. Kaapana develops on a private
GitLab and mirrors to GitHub, so its merged-pull-request count of zero
($\dagger$) is an artefact of mirroring. The mirrored history carries 206 merge
commits over the same twelve months, all of them in GitLab's \texttt{Merge
branch \dots{} into \dots} form rather than GitHub's. Similarly, MedPerf's only
tagged release ($\ddagger$) is marked as a pre-release, and the newest FLARE tag is a
release candidate (2.9.1rc1), so that column names 2.9.0, the last stable release. Third, stars measure
visibility rather than quality or fitness, and two of the most-starred projects in
the table have taken no commit in a year. Four platforms cannot be scored on
these metrics at all. GenoMed4All, INCISIVE and TrainTracks publish no
repository that we could locate, and PADME develops on a self-managed GitLab
whose API is not reachable by the method used here.

\section{Case Study: Researcher Journey on FLIP}
\label{sec:journey}

To ground the platform in practice, we describe the researcher's experience from project creation to the download of results.

A researcher with the appropriate role logs on to the FLIP web portal, creates a project, and writes a SQL cohort query against the OMOP schema. Each TN returns only aggregate counts for the matching cohort, with no row-level information. Figure~\ref{cohortsql} shows the number of records obtained per site in this study. The researcher refines the query until satisfied and then submits the project for approval, selecting which Trusts to include. The FLIP administrator of each selected site independently reviews the project and approves the site's participation, and the decision is recorded in the platform. Once the project is approved, each TN pulls the imaging cohort from PACS into its local XNAT instance. In this study, which used a simulated PACS server (Orthanc~\cite{Jodogne2026Orthanc}), image import took approximately 6~minutes per site for nearly 2,000 studies.

When the DICOM images have been imported, and optionally converted to NIfTI~\cite{cox2004nifti}, the researcher uploads the FL app files, which include the Python routines, model definition, configuration and, if needed, a pretrained checkpoint. FLIP scans uploaded files for vulnerabilities. Clicking ``Initiate Training'' dispatches the job to an available FL network, and progress can be monitored live through an activity feed and metric plots (Fig.~\ref{fig:flip-ui-screenshots}). The fine-tuning job executed 50 global rounds over $\approx$11.2~hours. At no point does the researcher need to access site infrastructure or execute commands on remote hosts. Upon completion, the downloadable results contain the aggregated model, per-round metrics, evaluation scores, and an audit trail. The evaluation job produces a structured JSON file with per-site, per-model AUROC values and the results of pairwise DeLong tests.

In summary, the researcher's effort consisted of writing one SQL query, adapting their code to FLIP following the available tutorials and documentation, uploading their FL app and running the job, while the platform handled all coordination, data access, governance, job scheduling, and model aggregation. This is the \emph{marginal} cost of a project at a site that has already been onboarded. One-time site onboarding is a separate and substantially larger effort, borne once per site rather than once per project. It covers TN installation, institutional security review (Section~\ref{subsec:governance}), harmonisation of local data into the OMOP schema, and PACS/XNAT integration.

\section{UK--Thailand CXR Fine-Tuning}
\label{sec:study}

We performed a cross-country federated study using fully synthetic CXR data, a setting chosen because CXR classifiers are well known to degrade when applied to data from external hospitals~\cite{yu2022,degrave2021,zech2018}. The UK cohort was generated from templated text prompts with GPT Image~2~\cite{openai2026gptimage2}, whereas the Thai cohort used image-conditioned generation. Prompts varied by target finding (pneumothorax, pleural effusion, consolidation, lung nodule/mass and infiltration) and laterality, producing 480 images per site per pathology at $768\times768$ pixels. Automated quality control (QC) with Gemini~3.5 Flash rejected 0.5\% of the UK and 2.5\% of the Thai images, and an expert radiographer reviewed a stratified subset. For each cohort, 20\% of the images were reserved for holdout evaluation, leaving $\approx$1,900 training images and $\approx$475 holdout images per site. We deployed two FL apps at both sites, one for federated training and one for federated evaluation.



\textbf{Federated Training Application.} We used Ark+~\cite{arkplus2026}, a CXR foundation model pretrained on six public datasets. The encoder was kept frozen, and a new 5-class linear head was trained from scratch with a binary cross-entropy loss. We used FedAvg with selective parameter sharing~\cite{shokri2015privacy} for 50 global rounds, with 5 local epochs per round. Under selective parameter sharing, only weight differences whose magnitude exceeds the $10\%$ quantile are shared, with absolute truncation at 0.01. Local training used a batch size of~4, with optimisers and learning rate schedulers persisting across global rounds. We used the Adam optimiser ($\beta_{1}{=}0.9$, $\beta_{2}{=}0.999$) with an initial learning rate of $1.5{\times}10^{-3}$, which decayed exponentially to $1.5{\times}10^{-4}$ during the first global round and remained fixed at this value thereafter.




\textbf{Federated Evaluation Application.} We implemented an application in which input models are packaged to a common inference interface, so that a single shared evaluator can run any of them without model-specific integration code. The evaluator executes the same protocol independently at each configured site. It first calculates per-class AUROC scores on the site's holdout set for all configured models. It then determines the best model for each class--site pairing following the significance-ranking protocol of the Medical Segmentation Decathlon~\cite{antonelli2022medical}, using a two-sided DeLong test with Benjamini--Hochberg false discovery rate (FDR) correction~\cite{benjamini1995} applied independently to each model pair at each site. We evaluated two models, each adapted to this interface. The first, \textbf{Pretrained Ark+}, is the original Ark+ model, with the outputs of its ChestX-ray14~\cite{wang2017chestx} classifier heads remapped to our five target classes. ``Nodule'' and ``Mass'' were combined into ``Nodule or Mass'' by taking the maximum of the two outputs, the other four target classes were taken directly from their corresponding heads, and the remaining upstream classes were discarded. The second, \textbf{Fine-tuned}, is Pretrained Ark+ after fine-tuning with our federated training application.

\subsection{System latency measurements}
FL deployments are paced by the slowest participant~\cite{fedscale2022}. One synchronous round decomposes as
\begin{equation}
	T_{\mathrm{round}} = T_{\mathrm{down}} + T_{\mathrm{compute}} + T_{\mathrm{up}} + T_{\mathrm{agg}} + T_{\mathrm{persist}} + T_{\mathrm{gap}},
	\label{eq:roundphases}
\end{equation}
that is, global-model download, local training, update upload, server-side aggregation, global-checkpoint persistence, and inter-round coordination. All terms are measured from the CH's own telemetry and recovered from the CH job export. Task dispatch and result receipt are timestamped per client in the audit trail, model transfers are bracketed by the object-transfer log, and aggregation, persistence and round boundaries are explicit server events. To isolate platform overhead from site compute, the same application was run in two further configurations. The first used the FLARE simulator at both sites, with the same FLARE controller and filter chain but no Wide Area Network (WAN), TLS, encryption or polling. The second was a deployed run with two clients inside one network, which retains the full platform stack but removes the WAN.

\section{Results}

\subsection{Model Performance}

Federated fine-tuning improved per-class AUROC in 6 of the 10 site--class pairings, and these improvements survived Benjamini--Hochberg correction at $Q{=}0.05$ (Table~\ref{tab:model-performance}). The exceptions were effusion at both sites, and consolidation and pneumothorax at the Thai site, where the pretrained model was already near ceiling ($\mathrm{AUROC} \geq 0.997$). Mean AUROC rose from 0.958 to 0.999 at the UK site (478 studies) and from 0.966 to 1.000 at the Thai site (468 studies). The fine-tuned model saturates both distributions ($\mathrm{AUROC} \geq 0.997$), which indicates that the synthetic task is easy, and these results should not be read as a state-of-the-art benchmark.

\begin{table}[!ht]
	\caption{Per-class AUROC of pretrained and FedAvg fine-tuned Ark+ on UK and Thai holdout sets. $p$-values are from two-sided DeLong tests, Benjamini--Hochberg (BH) corrected within each site. The testing family is the 5 classes for the single model pair at that site ($m{=}5$), so the BH threshold for the $k$-th smallest $p$-value is $0.05k/5$, ranging from $0.01$ to $0.05$. Four comparisons survive correction at the UK site and two at the Thai site, whereas effusion (both sites) and Thai consolidation and pneumothorax do not (n.s.). Bold values mark the best model per class and site at $Q{=}0.05$.}
	\label{tab:model-performance}
	\centering
	\footnotesize
	\setlength{\tabcolsep}{4pt}
	\begin{tabular}{@{}lcccccc@{}}
		\hline
		\textbf{Class} & \multicolumn{3}{c}{\textbf{UK Synthetic (n=478)}}
		               & \multicolumn{3}{c}{\textbf{Thai Synthetic (n=468)}}                                                                                             \\[0.4ex]
		\cline{2-4}\cline{5-7}
		               & \textbf{Pretr.}                                     & \textbf{F-tuned} & \textbf{$p$}
		               & \textbf{Pretr.}                                     & \textbf{F-tuned} & \textbf{$p$}                                                           \\[0.4ex]
		\hline
		Effusion       & 0.998                                               & 1.000            & 0.12 n.s.             & 1.000 & 1.000          & 1.00 n.s.             \\
		Consolidation  & 0.989                                               & \textbf{0.997}   & 0.027                 & 0.998 & 1.000          & 0.31 n.s.             \\
		Infiltration   & 0.863                                               & \textbf{0.998}   & $1.8{\times}10^{-14}$ & 0.878 & \textbf{1.000} & $1.2{\times}10^{-15}$ \\
		Nodule or Mass & 0.970                                               & \textbf{1.000}   & $5.9{\times}10^{-6}$  & 0.955 & \textbf{1.000} & $7.2{\times}10^{-7}$  \\
		Pneumothorax   & 0.971                                               & \textbf{1.000}   & 0.0065                & 0.997 & 1.000          & 0.19 n.s.             \\
		\hline
		Mean           & 0.958                                               & \textbf{0.999}   & ---                   & 0.966 & \textbf{1.000} & ---                   \\
		\hline
	\end{tabular}
\end{table}

\subsection{Latency Results}
\label{subsec:sysresults}

\begin{figure}[t]
	\centering
	\includegraphics[width=\textwidth]{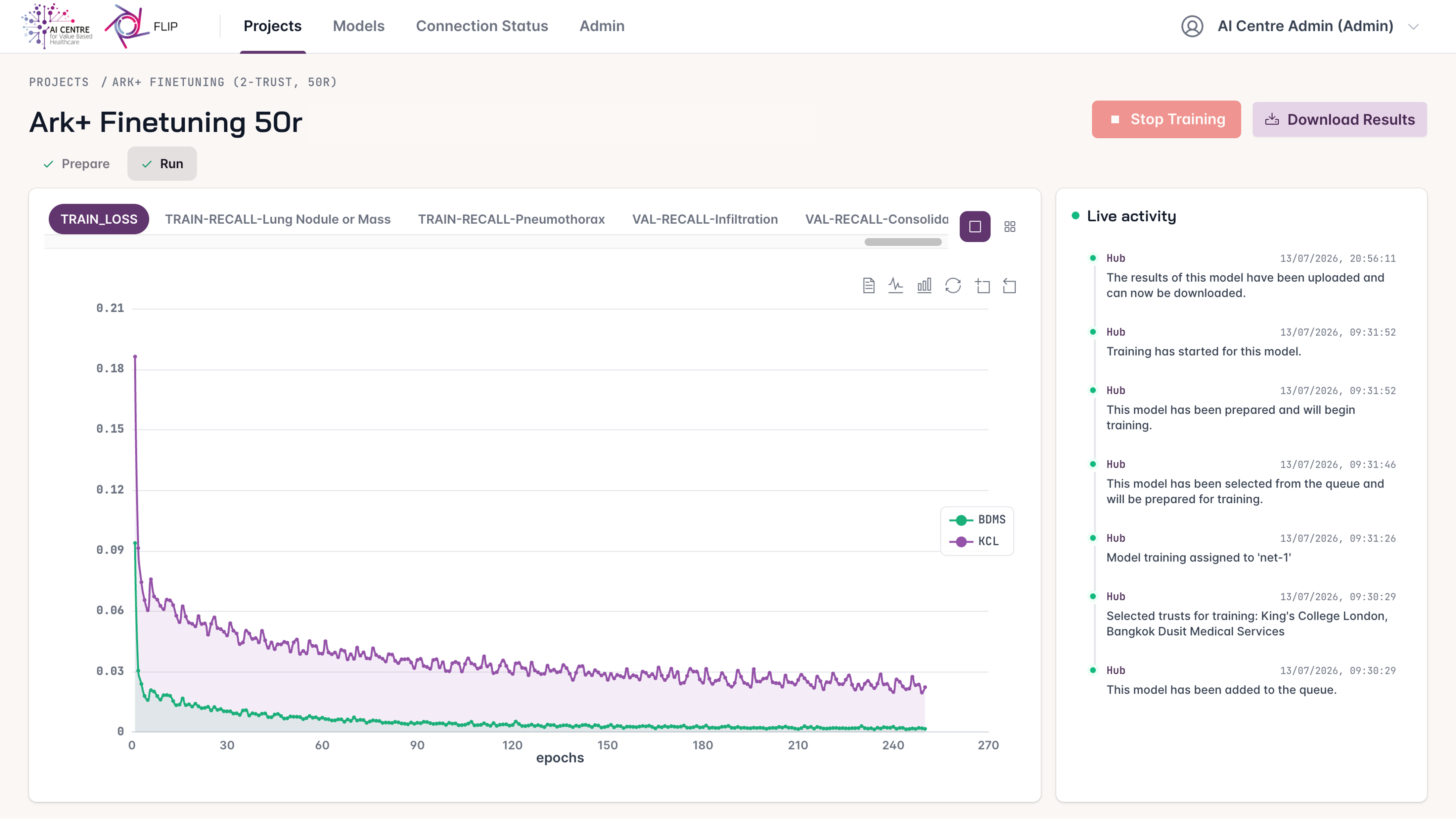}
	\caption{FLIP interface showing federated training results, UK--Thailand deployment.}
	\label{fig:flip-ui-screenshots}
\end{figure}

The 50-round federated run completed with no connection interruptions, retries, or client dropouts, with the FL server hosted in AWS eu-west-2 (Fig.~\ref{fig:flip-ui-screenshots}). Rounds 1--49 were stable at $745.8{\pm}0.7$~s. Server-side work comprised aggregation ($T_{\mathrm{agg}}{=}0.19{\pm}0.06$~s), global-checkpoint persistence ($T_{\mathrm{persist}}{=}5.52{\pm}0.04$~s) and inter-round coordination ($T_{\mathrm{gap}}{=}0.19{\pm}0.02$~s), which together took $5.9$~s, or $0.79\%$ of the round time (Table~\ref{tab:roundtiming}). Checkpoint persistence dominates this term by an order of magnitude, and coordination alone accounts for $0.05\%$.

The round is paced by the slower participant, and by compute rather than by the network. The Thai client's turnaround was $737.9{\pm}0.7$~s, or $98.9\%$ of the round, compared with $129.8{\pm}0.5$~s ($17.4\%$) at the UK site. Steady-state model transfer was negligible ($T_{\mathrm{down}}{<}0.2$~s), consistent with head-only updates of tens of kilobytes.

Round~0 took $3{,}830.0$~s, $3{,}084$~s longer than a steady-state round, and this excess is \emph{not} an effect of distance. It is dominated by one-time client-side initialisation at the UK site, whose round-0 turnaround exceeded its steady state by $3{,}698$~s. The UK site, rather than the cross-continental client, therefore paced round~0. The full-checkpoint broadcast to the Thai site added a further $954$~s over its steady state, which is a genuine but secondary cross-continental cost. The initialisation cost is incurred once and is amortised over a long run, but it would dominate short jobs and should be reported separately from the steady-state round cost.

\textbf{Reproducibility.} The same application runs without modification on FLIP's local simulator, with both clients on one GPU and no WAN or platform stack, and on the deployed platform. The simulator provides a baseline for cross-site replication without hospital infrastructure. The deployed platform preserves complete experiment provenance in the CH's database and object store, including the cohort SQL and site responses, uploaded application files and pretrained weights, per-round site metrics and logs, model-status and approval audit trails, and the final global model weights. A public demo page, the synthetic tutorial datasets, and the fine-tuned checkpoint are available at \url{https://app.flip.aicentre.co.uk/ark_demo}, \url{https://huggingface.co/datasets/aicentreflip/tutorials-arkplus-cxr-classification}, and \url{https://huggingface.co/aicentreflip/tutorials-arkplus-cxr-finetuned}, respectively. The release is pinned for reproduction. The FLIP source commit (\texttt{658c2304}), container image digests, deployment manifests, the full cohort SQL, both FL app bundles, random seeds, and complete experiment configurations are archived alongside it.

\begin{table}[!ht]
	\caption{Round-loop timing over 50 rounds (mean $\pm$ std, seconds; gap $n{=}49$). Each simulator serialises both clients on one GPU, so its round time is approximately the sum of both sites' $T_{\mathrm{compute}}$. Client turnaround runs from task dispatch to acceptance of the contribution, and therefore spans $T_{\mathrm{down}}{+}T_{\mathrm{compute}}{+}T_{\mathrm{up}}$. Dashes mark phases not yet re-derived from the corresponding job export, as only the UK--TH deployment has been decomposed at this granularity.}
	\label{tab:roundtiming}
	\centering
	\small
	\begin{tabular*}{\linewidth}{@{\extracolsep{\fill}}lrrrr}
		\hline
		\textbf{Phase} & \textbf{Depl.\ UK--TH} & \textbf{Depl.\ same-net} &
		\textbf{Sim.\ UK} & \textbf{Sim.\ TH} \\
		\hline
		Rounds 1--49 ($T_{\mathrm{round}}$) & $745.8 \pm 0.7$ & $181.7 \pm 0.8$ &
		$275.4 \pm 0.5$ & $1{,}605.2 \pm 2.3$ \\
		Round 0 (init.)                     & $3{,}830.0$     & $334.6$         &
		$294.7$         & $1{,}696.2$ \\
		Aggregation ($T_{\mathrm{agg}}$)    & $0.19 \pm 0.06$ & $0.18 \pm 0.04$ &
		$0.22 \pm 0.05$ & $0.39 \pm 0.09$ \\
		Checkpoint persist.\ ($T_{\mathrm{persist}}$) & $5.52 \pm 0.04$ & --- & --- & --- \\
		Inter-round gap ($T_{\mathrm{gap}}$)& $0.19 \pm 0.02$ & $0.19 \pm 0.02$ &
		$0.21 \pm 0.02$ & $0.29 \pm 0.03$ \\
		Model transfer ($T_{\mathrm{down}}$)& $<0.2$          & ---             & --- & --- \\
		\hline
		Client turnaround, UK               & $129.8 \pm 0.5$ & ---             & --- & --- \\
		Client turnaround, Thailand         & $737.9 \pm 0.7$ & ---             & --- & --- \\
		\hline
	\end{tabular*}
\end{table}

\section{Discussion and Conclusion}
This study demonstrates an open, multi-application platform for repeatable FL. Our main contribution is operational rather than algorithmic. Two distinct application types, fine-tuning and evaluation, ran across the UK and Thailand sites through a single platform, governed by independent site-level approvals.

\textbf{Generality.} FLIP was exercised across two distinct workloads (federated CXR fine-tuning and federated model evaluation) using the same service architecture, cohort-query mechanism, and job-type system without platform-level changes. The only project-specific artefacts were the SQL query and application files; the platform handled scheduling, data access, aggregation, and provenance.

\textbf{Researcher experience.} The researcher's journey, which consisted of writing one query, uploading files and clicking ``Initiate Training'', required no access to site infrastructure, no remote command execution, and no platform reconfiguration between jobs. The architecture maps onto the operational concerns of a multi-institutional study, namely cohort definition (OMOP query), data access (imaging API), execution (FL client on GPU), governance (per-site approval), and provenance (audit trails). Each concern is owned by the appropriate party, and the researcher interacts only with the CH.

\textbf{Cohort flexibility.} Any cohort definable in OMOP (by procedure, diagnosis, demographics, or laboratory values) is supported, provided that the data have been harmonised and ingested into each site's database. The platform's imaging API resolves matching DICOM series on demand.

\textbf{What this study does not claim.} No centralised comparator was trained, and none was intended. The deployment used synthetic cohorts on a task that the fine-tuned model saturates ($\mathrm{AUROC} \geq 0.997$). Table~\ref{tab:model-performance} therefore demonstrates that the federation executed correctly and that the platform carried a real fine-tuning workload across two continents, but not that federated training is preferable to pooling. The latency measurements are similarly limited, as they characterise one 50-round job over two nodes rather than the cost of federation in general. Readers assessing whether federated training is accurate enough for a particular clinical task should consult the aggregate evidence cited in Section~\ref{sec:intro} rather than this paper, which was designed to answer an operational question and is not powered to answer a statistical one. FLIP is premised on the accuracy cost of federation being real, modest and worth paying. The constraint that rules out pooling is legal and institutional rather than technical, and the practical alternative to a federated study is usually a single-site study rather than a pooled one. Our contribution is therefore to show that the governance and integration work that normally makes such collaborations one-off can be made reusable.

\textbf{Limitations.} As a proof of concept, this deployment exercised one of FLIP's two supported FL backends (FLARE), one aggregation strategy (FedAvg), one institution in each country, two applications (federated fine-tuning and federated evaluation), and governance for a synthetic-data project only. Broader backend coverage, larger federations, and governance under real patient data remain untested. The example applications of Section~\ref{subsec:generalPurpose} were not deployed across the two sites. Some of the remaining limitations concern the deployment rather than the platform's capabilities. The deployment ran an earlier release, in which approver identity was not yet trust-scoped (Section~\ref{subsec:governance}) and runtime policy was not yet declarative, and it therefore did not exercise the mechanisms added in release~\flipversion. One limitation of the design itself remains, namely that provenance is emitted as a model-bundle record that the hub does not attest.

\textbf{Future work.} Any use case whose cohort can be defined in OMOP and DICOM, and whose experiments can be packaged as a FLARE or Flower app, is a natural candidate for FLIP. As FLIP extends to more sites and heterogeneous hardware~\cite{Li2020FLChallenges}, future work will need to address non-IID data heterogeneity~\cite{noniidsurvey2024}, real clinical data under governance approvals, and further job types, of which the current catalogue covers segmentation, classification, evaluation and synthesis but not detection.

\section*{Competing Interests}
The authors have no competing interests to declare that are relevant to the content of this article.

%
%
\bibliographystyle{unsrturl}
\bibliography{mybibliography}


\appendix
\section{Search Strings}
\label{app:search}

The four searches below are reproduced exactly as executed. The PubMed and
GitHub queries are the machine-readable content of \texttt{search\_protocol.sh},
the arXiv query is generated by \texttt{fetch\_arxiv.py}, and the medRxiv sweep
by \texttt{fetch\_medrxiv.py}. All three scripts are released with the paper, so
readers can re-run the identification rather than take the counts on trust.
Counts in Sect.~\ref{subsec:inclusion} are from the run of 5~September 2026. A
re-run on 10~September 2026 returned 1{,}485 records, 2{,}330 repositories,
1{,}190 arXiv preprints and 167 medRxiv matches, and the difference reflects new
deposits rather than a change of query.

\subsection{PubMed}
\label{app:pubmed}

The query is the conjunction of three concept blocks, \texttt{A AND B AND C}.
Block~A is load-bearing in its free-text form: the MeSH descriptor
\emph{Federated Learning} indexes no record dated before 2024, and the
\emph{federated evaluation} and \emph{federated benchmarking} phrases are what
retrieve MedPerf, whose title claims neither federated learning nor a platform.

\vspace{0.5em}
\noindent\textbf{Block A: federated computation}
\begin{verbatim}
("Federated Learning"[MeSH]
 OR "federated learning"[tiab]
 OR "federated analytics"[tiab]
 OR "federated evaluation"[tiab]
 OR "federated benchmarking"[tiab]
 OR "federated training"[tiab]
 OR "federated infrastructure"[tiab]
 OR "swarm learning"[tiab]
 OR "personal health train"[tiab]
 OR "distributed analysis"[tiab])
\end{verbatim}

\noindent\textbf{Block B: a system rather than a study}
\begin{verbatim}
(platform*[tiab] OR framework*[tiab]
 OR infrastructur*[tiab] OR toolkit*[tiab]
 OR software[tiab] OR "open source"[tiab]
 OR open-source[tiab] OR system*[tiab] OR tool*[tiab])
\end{verbatim}

\noindent\textbf{Block C: health setting}
\begin{verbatim}
(clinic*[tiab] OR hospital*[tiab] OR patient*[tiab]
 OR medical[tiab] OR health*[tiab] OR biomedical[tiab]
 OR imaging[tiab] OR radiolog*[tiab] OR multicent*[tiab]
 OR multi-cent*[tiab])
\end{verbatim}

\subsection{GitHub}
\label{app:github}

The GitHub code-search API rejects any query using more than five boolean
operators, so the strategy is a union of eighteen narrow queries rather than one
expression; the union is deduplicated by full repository name. Every query
carries \texttt{archived:false fork:false}.

No query applies a star floor. An earlier version of this protocol restricted
three of the queries to repositories with at least 20 or 50 stars, which was the
wrong instrument for this subject. The engines in this comparison hold a median
of 903 stars, against 110 for the deployed healthcare platforms and 32 for the
analytics infrastructures, so any star floor removes the clinically deployed
platforms first. However, removing the floor alone only grows the pool and
recovers none of the missed comparators, because the binding constraint was
never popularity but query vocabulary. Repositories that carry no matching topic
tag and no matching description phrase (vantage6, \texttt{dsBase} and
FeatureCloud) are invisible to a topic-and-keyword query at any threshold, and
must be named explicitly. The last eight queries below are these name-targeted
additions.

In place of the star floor, we apply a contributor floor as a post-fetch filter,
retaining a repository only if the contributors endpoint reports at least one
contributor. GitHub's search API exposes no collaborator or contributor
qualifier (\texttt{users:>=n} returns nothing and \texttt{contributors:>=n} is
silently ignored), so the filter cannot be expressed in the query and is applied
to the fetched union instead. At this threshold, the filter acts mainly as a
liveness check. Of the 2{,}432 repositories returned by the queries, it retains
2{,}330 and removes 102 that no longer resolve or expose no contributors.
Raising the floor to two would additionally remove the approximately 2{,}100
single-maintainer repositories that dominate the raw pool, but it would also
remove FLA\textsuperscript{3}, which has a single author. We therefore keep the
floor at one and rely on the inclusion criteria for the remaining selection.

\begin{verbatim}
topic:federated-learning pushed:>=2024-09-01
topic:federated-learning-framework
topic:federated topic:healthcare
topic:federated-learning topic:healthcare
topic:federated-learning topic:medical-imaging
topic:federated-analytics
"federated learning" in:name,description
"federated evaluation" in:name,description
"personal health train" in:name,description,readme
"swarm learning" in:name,description
vantage6 in:name,description
dsBase in:name
FeatureCloud in:name
FL4Med in:name
GenoMed4All in:name,description
nebula federated in:name,description
FLAAA in:name
TrainTracks in:name,description
\end{verbatim}

\subsection{arXiv}
\label{app:arxiv}

The arXiv query is the same three-concept conjunction as PubMed, expressed in
arXiv's search syntax. Two of its constraints follow from arXiv's design
rather than from our choice: there is no MeSH, so Block~A is free text
throughout; and there is no stemming or wildcard, so Block~B and Block~C list
singular and plural forms explicitly. \texttt{ti:} and \texttt{abs:} are
arXiv's title and abstract fields, standing in for PubMed's \texttt{[tiab]}.
The query returned 1{,}190 records. Because a bare arXiv record is a preprint,
it is counted as satisfying criterion~(iv) only when its metadata also carries
a public repository link or a journal reference.

\begin{verbatim}
((ti:"federated learning"
  OR abs:"federated learning")
 OR (ti:"federated analytics"
  OR abs:"federated analytics")
 OR (ti:"federated evaluation"
  OR abs:"federated evaluation")
 OR (ti:"federated benchmarking"
  OR abs:"federated benchmarking")
 OR (ti:"federated training"
  OR abs:"federated training")
 OR (ti:"federated infrastructure"
  OR abs:"federated infrastructure")
 OR (ti:"swarm learning" OR abs:"swarm learning")
 OR (ti:"personal health train"
  OR abs:"personal health train")
 OR (ti:"distributed analysis"
  OR abs:"distributed analysis"))
AND ((ti:"platform" OR abs:"platform")
 OR (ti:"platforms" OR abs:"platforms")
 OR (ti:"framework" OR abs:"framework")
 OR (ti:"frameworks" OR abs:"frameworks")
 OR (ti:"infrastructure" OR abs:"infrastructure")
 OR (ti:"toolkit" OR abs:"toolkit")
 OR (ti:"toolkits" OR abs:"toolkits")
 OR (ti:"software" OR abs:"software")
 OR (ti:"open source" OR abs:"open source")
 OR (ti:"system" OR abs:"system")
 OR (ti:"systems" OR abs:"systems")
 OR (ti:"tool" OR abs:"tool")
 OR (ti:"tools" OR abs:"tools"))
AND ((ti:"clinic" OR abs:"clinic")
 OR (ti:"clinical" OR abs:"clinical")
 OR (ti:"hospital" OR abs:"hospital")
 OR (ti:"hospitals" OR abs:"hospitals")
 OR (ti:"patient" OR abs:"patient")
 OR (ti:"patients" OR abs:"patients")
 OR (ti:"medical" OR abs:"medical")
 OR (ti:"health" OR abs:"health")
 OR (ti:"healthcare" OR abs:"healthcare")
 OR (ti:"biomedical" OR abs:"biomedical")
 OR (ti:"imaging" OR abs:"imaging")
 OR (ti:"radiology" OR abs:"radiology")
 OR (ti:"multicenter" OR abs:"multicenter")
 OR (ti:"multi-center" OR abs:"multi-center")
 OR (ti:"multi-institutional"
  OR abs:"multi-institutional"))
\end{verbatim}

\subsection{medRxiv}
\label{app:medrxiv}

The medRxiv/bioRxiv API has no search endpoint and serves metadata only by date
range with cursor pagination, so the query cannot be submitted to the server.
Instead, it is applied locally. All 109{,}695 records in the window from
medRxiv's launch (June 2019) to the review cutoff are fetched, and each title
and abstract is matched against the same two free-text blocks. Block~C is
omitted because medRxiv is itself a health preprint server, so every record
satisfies it by construction. The sweep returned 167 matches. Because a medRxiv
record is a preprint, it satisfies criterion~(iv) only when it also carries a
\texttt{published} field (a later journal DOI) or names a public repository.

\begin{verbatim}
A  federat* OR "swarm learning" OR "personal health train"
   OR "distributed analysis" OR "split learning"
B  platform* OR framework* OR infrastructure* OR toolkit*
   OR software OR "open source" OR system* OR tool*
   (C omitted: the server is medical by construction)
matched = A AND B, over 109,695 records, 2019-06-01..2026-09-05
\end{verbatim}

\subsection{What the queries do not retrieve}
\label{app:recall}

Table~\ref{tab:recall} records, for every comparator, whether the pool generated
by each query set actually contained it. The four searches complement each other
in a structured way. PubMed recovers the analytics layer but misses engines,
which publish in computer-science venues that it does not index. GitHub recovers
the engines completely and, once the query set names them explicitly, the
repository-bearing platforms as well, missing only FedKBP\textsuperscript{+},
whose repository is an empty placeholder, and gLinDA, which is hosted on a
university GitLab rather than on GitHub. arXiv shares PubMed's health
restriction and therefore also misses the engines. Finally, medRxiv recovers
the one comparator that the other three miss, GenoMed4All, whose only public
description is a medRxiv preprint that PubMed does not index until journal
publication and arXiv does not host. Every comparator is therefore retrieved by
at least one of the four searches, and none depends on citation tracking alone.
The medRxiv sweep also surfaces two platforms absent from the other three pools,
the Biomedical Research Hub and Cumulus. These are not tabulated because their
published form is a federated \emph{analytics} or data-sharing system rather
than a federated learning platform.

\begin{table}[!ht]
	\caption{Retrieval of each comparator by the four searches, measured against
		the saved candidate pools. \emph{yes} the pool contained an artefact for
		that platform; \emph{no} it did not; \emph{--} the platform has no artefact
		of that kind to retrieve (no indexed paper, or no public repository). The
		PubMed and arXiv columns are assessed over the 1{,}485-record and
		1{,}190-preprint pools respectively, the GitHub column over the
		2{,}330-repository union, and the medRxiv column over the 167-preprint match
		pool. A platform paper here means a record describing the platform, not a
		study merely citing it.}
	\label{tab:recall}
	\centering
	\scriptsize
	\setlength{\tabcolsep}{3.5pt}
	\renewcommand{\arraystretch}{1.1}
	\begin{tabular}{@{}>{\raggedright\arraybackslash}p{2.9cm}>{\raggedright\arraybackslash}p{2.2cm}cccc@{}}
		\hline
		\textbf{Platform} & \textbf{Layer} & \textbf{PubMed} & \textbf{GitHub} & \textbf{arXiv} & \textbf{medRxiv} \\
		\hline
		APPFL                  & deployed  & yes & yes & no  & no  \\
		CODA                   & deployed  & yes & --  & no  & no  \\
		FLA\textsuperscript{3} & deployed  & --  & yes & yes & no  \\
		FeatureCloud           & deployed  & --  & yes & yes & no  \\
		FeTS                   & deployed  & yes & yes & yes & no  \\
		FL4Health              & deployed  & --  & yes & no  & no  \\
		GenoMed4All            & deployed  & --  & --  & no  & yes \\
		INCISIVE               & deployed  & yes & --  & no  & no  \\
		JIP/Kaapana            & deployed  & --  & yes & no  & no  \\
		MedPerf                & deployed  & yes & yes & yes & no  \\
		PHT-meDIC/PADME        & deployed  & yes & yes & yes & yes \\
		\hline
		Swarm Learning         & p2p       & --  & yes & yes & yes \\
		NEBULA/Fedstellar      & p2p       & --  & yes & no  & no  \\
		FedKBP\textsuperscript{+} & p2p    & yes & no  & yes & no  \\
		\hline
		vantage6               & analytics & yes & yes & no  & yes \\
		TrainTracks            & analytics & yes & --  & no  & no  \\
		DataSHIELD             & analytics & yes & yes & yes & yes \\
		gLinDA                 & analytics & yes & no  & no  & no  \\
		\hline
		FLARE                  & engine    & yes & yes & yes & no  \\
		Flower                 & engine    & --  & yes & yes & no  \\
		OpenFL                 & engine    & yes & yes & yes & no  \\
		Fed-BioMed             & engine    & --  & yes & yes & no  \\
		Substra                & engine    & --  & yes & yes & no  \\
		FEDn                   & engine    & --  & yes & no  & no  \\
		FATE                   & engine    & --  & yes & no  & no  \\
		FedML                  & engine    & --  & yes & yes & no  \\
		\hline
		\multicolumn{2}{@{}l}{\emph{Retrieved by at least one search}}
		                                   & \multicolumn{4}{c}{26 of 26} \\
		\hline
	\end{tabular}
\end{table}

\end{document}